\documentclass[10pt,twocolumn]{article}

\usepackage[letterpaper,margin=0.75in,headheight=14pt,columnsep=0.25in]{geometry}
\usepackage{lmodern}
\usepackage[T1]{fontenc}
\usepackage[utf8]{inputenc}
\usepackage{microtype}
\usepackage{setspace}
\usepackage{fancyhdr}
\usepackage{lastpage}

\makeatletter
\renewcommand{\maketitle}{%
  \begingroup
    \centering
    {\LARGE\bfseries \@title\par}%
    \vspace{1.2em}%
    {\large \@author\par}%
    \vspace{0.6em}%
    {\small \@date\par}%
    \vspace{1.5em}%
  \endgroup
  \thispagestyle{fancy}%
}

\@addtoreset{section}{part}
\renewcommand\section{\@startsection{section}{1}{\z@}%
  {-3.5ex \@plus -1ex \@minus -.2ex}%
  {2.3ex \@plus.2ex}%
  {\normalfont\Large\bfseries}}
\renewcommand\subsection{\@startsection{subsection}{2}{\z@}%
  {-3.25ex\@plus -1ex \@minus -.2ex}%
  {1.5ex \@plus .2ex}%
  {\normalfont\large\bfseries}}
\renewcommand\subsubsection{\@startsection{subsubsection}{3}{\z@}%
  {-3.25ex\@plus -1ex \@minus -.2ex}%
  {1.5ex \@plus .2ex}%
  {\normalfont\normalsize\bfseries}}
\makeatother
\usepackage[htt]{hyphenat}
\makeatletter
\@ifundefined{l@nohyphenation}{\newlanguage\l@nohyphenation}{}
\let\xlog@texttt\texttt
\renewcommand{\texttt}[1]{{\language\l@nohyphenation
  \xlog@texttt{\hyphenchar\font\m@ne\relax #1}}}
\renewcommand{\_}{\textunderscore\hspace{0pt}}
\makeatother
\usepackage{amsmath,amssymb}
\usepackage{algorithm}
\usepackage[noend]{algpseudocode}
\usepackage{etoolbox}

\makeatletter
\algrenewcommand\ALG@beginalgorithmic{\footnotesize}
\makeatother
\algrenewcommand\alglinenumber[1]{{\scriptsize\color{figgray}#1:}}
\algrenewcommand{\algorithmiccomment}[1]{%
  \hfill{\scriptsize\itshape\color{figgray}$\triangleright$\ #1}}

\usepackage{booktabs}
\usepackage{tabularx}
\usepackage{array}
\newcolumntype{L}{>{\raggedright\arraybackslash}X}
\usepackage{graphicx}
\usepackage{tikz}
\usetikzlibrary{arrows.meta,positioning,fit,backgrounds,calc,shapes.geometric}
\usepackage{listings}
\usepackage{xcolor}
\usepackage{caption}
\usepackage[hidelinks,colorlinks=true,
            linkcolor=black!70!blue,
            citecolor=black!70!blue,
            urlcolor=black!70!blue]{hyperref}
\usepackage[noabbrev,capitalise]{cleveref}
\crefname{algorithm}{Algorithm}{Algorithms}
\Crefname{algorithm}{Algorithm}{Algorithms}
\usepackage[backend=biber,style=numeric-comp,sorting=none]{biblatex}
\graphicspath{{./}}

\definecolor{codebg}{RGB}{248,248,248}
\definecolor{codekey}{RGB}{0,90,156}
\definecolor{codecomment}{RGB}{106,115,125}
\definecolor{codestring}{RGB}{163,21,21}

\lstdefinelanguage{xlog}{
  morekeywords={pred,func,domain,use,private,nn,true,false,not,and,or,if,then,else,is,query,fact,rule,count,sum,min,max,logsumexp,abs,pow,cast},
  sensitive=true,
  morecomment=[l]{//},
  morecomment=[s]{/*}{*/},
  morestring=[b]",
  literate={:-}{{{\color{codekey}:-}}}2
           {::}{{{\color{codekey}::}}}2
           {?-}{{{\color{codekey}?-}}}2
           {->}{{{\color{codekey}->}}}2
}
\lstdefinestyle{codebase}{
  backgroundcolor=\color{codebg},
  basicstyle=\ttfamily\footnotesize,
  keywordstyle=\color{codekey}\bfseries,
  commentstyle=\color{codecomment}\itshape,
  stringstyle=\color{codestring},
  frame=single,
  framerule=0.3pt,
  breaklines=true,
  breakatwhitespace=true,
  breakindent=0pt,
  prebreak=\mbox{\textcolor{codecomment}{$\hookleftarrow$}},
  postbreak=\mbox{\textcolor{codecomment}{$\hookrightarrow$}\space},
  columns=fullflexible,
  keepspaces=true,
  showstringspaces=false,
  numbers=none,
  captionpos=b,
  abovecaptionskip=0.5em,
  belowcaptionskip=0.5em,
}
\lstdefinestyle{xlogstyle}{style=codebase,language=xlog}
\lstdefinestyle{pythonstyle}{style=codebase,language=Python}
\crefname{lstlisting}{Listing}{Listings}
\Crefname{lstlisting}{Listing}{Listings}

\BeforeBeginEnvironment{lstlisting}{%
  \par\medskip\noindent\begin{minipage}{\linewidth}}
\AfterEndEnvironment{lstlisting}{\end{minipage}\par\medskip}
\let\xlogInputListing\lstinputlisting
\renewcommand{\lstinputlisting}[2][]{%
  \par\medskip\noindent\begin{minipage}{\linewidth}%
  \xlogInputListing[#1]{#2}%
  \end{minipage}\par\medskip}

\definecolor{figblue}{RGB}{0,90,156}
\definecolor{figgreen}{RGB}{27,128,79}
\definecolor{figgray}{RGB}{90,98,110}
\definecolor{figink}{RGB}{45,52,64}
\definecolor{figband}{RGB}{235,242,249}
\definecolor{figbandedge}{RGB}{170,197,222}
\definecolor{figaccent}{RGB}{176,42,42}
\definecolor{figchip}{RGB}{243,246,249}
\tikzset{
  fnode/.style={draw=figgray,rounded corners=2pt,fill=white,align=center,
                font=\small,inner sep=4pt,minimum height=8mm},
  fproc/.style={fnode,draw=figblue,line width=0.9pt},
  fcert/.style={fnode,draw=figaccent,line width=0.9pt},
  fneural/.style={fnode,draw=figgreen,line width=0.9pt},
  flabel/.style={font=\scriptsize\itshape,text=figgray,align=center},
  fsub/.style={font=\scriptsize,text=figgray},
  ftier/.style={draw=figink,fill=white,rounded corners=2pt,align=center,
                inner sep=3pt},
  fplane/.style={fill=figband,draw=figbandedge,rounded corners=3pt,
                 line width=0.5pt},
  fplanetitle/.style={fill=figband,draw=figbandedge,rounded corners=1.6pt,
                      line width=0.5pt,font=\scriptsize\scshape,
                      text=figblue,inner xsep=4pt,inner ysep=2.4pt},
  farrow/.style={-{Stealth[length=2.2mm]},figink,line width=0.8pt},
  fback/.style={-{Stealth[length=2.2mm]},figgreen,line width=0.8pt,dashed},
  fver/.style={-{Stealth[length=1.8mm]},figaccent,line width=0.7pt},
  pics/gpuchip/.style={code={%
    \foreach \p in {-0.75,0,0.75}{%
      \draw[figblue,line width=0.45pt] (\p mm,1.15mm)--(\p mm,1.7mm)
                                       (\p mm,-1.15mm)--(\p mm,-1.7mm)
                                       (1.15mm,\p mm)--(1.7mm,\p mm)
                                       (-1.15mm,\p mm)--(-1.7mm,\p mm);}%
    \draw[figblue,line width=0.6pt,fill=white]
      (-1.15mm,-1.15mm) rectangle (1.15mm,1.15mm);
    \draw[figblue,line width=0.4pt,fill=figblue!20]
      (-0.5mm,-0.5mm) rectangle (0.5mm,0.5mm);
  }},
}
\newcommand{\cratechip}[1]{{\setlength{\fboxsep}{1.6pt}\setlength{\fboxrule}{0.4pt}%
  \fcolorbox{figgray!45}{figchip}{\scriptsize\ttfamily #1}}}
\newcommand{\planetitle}[1]{\tikz[baseline=-0.6ex]{\pic{gpuchip};}\hspace{1.4mm}#1}
\newcommand{\fsubt}[1]{{\scriptsize\color{figgray}#1}}

\title{{\Huge\bfseries XLOG}\\[0.45em]
A CUDA-Native Engine for Neurosymbolic Integration}
\author{Levi Dubrovin, Nikita Pospelov, Kirill Sabitov\\[0.35em]\small Brainyblaze Dynamics Inc.\\\small \texttt{levi@brainyblaze.com}, \texttt{nikita\_pospelov@brainyblaze.com}, \texttt{kirill@brainyblaze.com}\\[0.35em]\small \url{https://github.com/BrainyBlaze/xlog}\\[0.35em]\small Technical Whitepaper}
\date{July 2026}

\begin{document}
\twocolumn[
  \begin{@twocolumnfalse}
    \maketitle
    \begin{abstract}
\noindent
xlog is a CUDA-native logic programming engine that couples neural perception with deterministic Datalog, probabilistic inference, and epistemic reasoning over world views through a shared typed frontend and provider-owned CUDA runtime. The routes share device data planes but not one universal execution boundary: ordinary Datalog and exact inference are host-orchestrated, while the certified resident recursive and Monte Carlo sampled cores record zero tracked host--device transfers and then return one bounded terminal receipt. End-to-end gradients are realized through the probabilistic path. A CUDA-backed knowledge-compilation pipeline (provenance $\to$ CNF $\to$ Decision-DNNF $\to$ circuit) turns neural outputs into a differentiable circuit whose forward pass is exact weighted model counting and whose backward pass yields exact gradients. The final smoothed circuit is certified against its back-end source formula before it is cached or evaluated: bounded status and error scalars are observed by the host, while resolution-proof traces are checked on the GPU. A resident Monte Carlo alternative covers supported programs for which approximate inference is selected. Circuit caching yields a $2.74\times$ MNIST-addition training speedup, and a worst-case-optimal join subsystem a $27.96\times$ geometric-mean gain over xlog's own binary-join baseline. Against external engines the picture is mixed: MNIST-addition accuracy matches Scallop's ($0.9561$ against $0.9468$) and we make no per-epoch speed claim, because the baseline's epoch time is not monotone in the host's CPU quota; across five hub-skewed triangle-counting cases the Soufflé-to-fused-xlog execution-time ratio \emph{grows with the input}, from $0.88\times$ at $150$k edges --- where Soufflé is the faster engine --- to $5.54\times$ at $1.2$M, while fused peak device allocations stay between $85$ and $1{,}033$\,MB against $3{,}287$ to $44{,}979$\,MB for the materializing arm, and every arm completes; and exact inference is correctness-equivalent to but slower than ProbLog2. On a public video benchmark, a proximity predicate trained through symbolic credit alone, never shown a distance label, replaces hand-set geometry at no cost in held-out accuracy; the same detector inside the Event-Calculus rule search does not survive ten-fold cross-validation, and on a leak-free split the committed clause does not transfer. On a maritime corpus of thousands of gold events, weighted clauses beat crisp selection by $0.065$ F1, and a single chronological training pass reproduces that result exactly.
\end{abstract}

    \vspace{1.5em}
  \end{@twocolumnfalse}
]

\section{Introduction}\label{sec:intro}

Neurosymbolic systems combine neural perception with symbolic reasoning, but established integrations often place the two halves on different processors. Deep learning frameworks---PyTorch, JAX---execute dense tensor computation on the GPU through optimized CUDA kernels, while systems such as DeepProbLog~\cite{deepproblog} and NeurASP~\cite{neurasp} use CPU-side symbolic backends. Their integration can therefore cross the PCIe bus for network outputs, symbolic results, or gradients. At scale, repeated host--device movement can dominate wall-clock time and bandwidth.

The transfer wall is one half of the problem; fragmented interfaces are the other. Existing neurosymbolic frameworks commonly bridge a neural network to one symbolic backend (DeepProbLog to probabilistic logic, NeurASP to answer-set programming). GPU logic efforts, conversely, accelerate focused workloads such as deterministic Datalog~\cite{gpulog,vflog} or SAT~\cite{parafrost}. xlog explores how a typed language and shared CUDA services can support deterministic, probabilistic, epistemic, and neural-symbolic use without pretending that those routes have identical execution or differentiability properties.

xlog is a CUDA-native logic programming language whose compiler and runtime treat a neural network as an ordinary predicate. Network outputs can become weighted facts for probabilistic knowledge compilation, where exact query gradients flow back to the network. Other reasoning modes reuse the typed relational and solver interfaces without claiming the same gradient semantics. Device-resident buffers carry facts, deltas, circuits, solver state, and gradients, while the host performs orchestration, I/O, compilation, and bounded control reads. The certified resident recursive and Monte Carlo sampled cores provide the stronger measured boundary: no tracked host--device transfers inside the core, followed by one terminal receipt. Verified knowledge compilation certifies the final smoothed circuit against its back-end formula before use, and DLPack~\cite{dlpack} exports device gradient buffers directly into PyTorch's autograd graph.

The principal contributions of this paper are:

\begin{itemize}
  \item \textbf{A shared neurosymbolic integration model.} A neural network is a predicate, and typed device-buffer interfaces connect it to the supported reasoning routes. End-to-end gradient training is realized specifically through the probabilistic knowledge-compilation path (\Cref{sec:arch,sec:nesy}).
  \item \textbf{CUDA relational execution with a certified resident route.} Semi-naive Datalog evaluation with stratified negation and aggregation uses custom CUDA relational kernels. Eligible recursive query plans are selected automatically for a resident conditional graph whose measured core has no tracked transfers; the wider ordinary executor remains host-orchestrated. A worst-case-optimal join (WCOJ) subsystem avoids the intermediate-relation blowup of binary-join plans on skewed cyclic queries (a measured $27.96\times$ geometric-mean ablation, \Cref{sec:datalog}).
  \item \textbf{Verified knowledge compilation.} A CUDA-backed provenance $\to$ CNF $\to$ Decision-DNNF $\to$ circuit pipeline certifies the final smoothed circuit against its back-end source formula before caching or evaluation. The host observes bounded solver status and error scalars; resolution-proof traces are checked on the GPU. Structural invariants required for weighted model counting are established separately by construction (\Cref{sec:prob}).
  \item \textbf{End-to-end differentiable training with zero-copy interop.} Compiled circuits are cached across iterations and evaluated as level-parallel forward/backward kernels, with gradients exported zero-copy via DLPack and Arrow; circuit caching yields a measured $2.74\times$ training speedup. Term embeddings and differentiable ILP ride the same device-resident path (\Cref{sec:nesy}).
  \item \textbf{Perception learned through symbolic credit on an external benchmark.} On the CAVIAR video-event corpus, under a direct per-timestep target, the rule search returns the same theory whether its proximity predicate is the hand-set geometric one that dedicated rule learners are handed or a network trained by the logic's own credit alone, never shown a distance label or any target of its own, at no cost in held-out accuracy. Event-Calculus rules over the precomputed predicate are induced under pre-registered gates; substituting the learned detector inside \emph{that} search is where the result currently fails (\Cref{sec:ecinduction}).
  \item \textbf{Rule induction at scale.} On the Brest AIS maritime corpus (3{,}548 gold \texttt{rendezVous} intervals) the same protocol, over pair-level folds and permutation-null gates, measures in isolation the two mechanisms the published Event-Calculus learners are built on. Weighting the clauses of the same gated pool raises micro F1 by $0.065$ over crisp selection, against our own prior expectation that it would not, and learning those weights in a single chronological pass reproduces the batch predictions exactly, for a reason we can name. We draw no comparison with the published maritime figure, whose temporal grid differs from ours (\Cref{sec:maritime}).
  \item \textbf{Reasoning beyond probability.} Epistemic reasoning over world views (\texttt{know}/\texttt{possible} under founded and Gelfond-style semantics) and well-founded semantics execute on the same substrate over finite supported programs, including positive modal fixpoints and supported recursion through negation evaluated by GPU-backed well-founded semantics, reusing the CDCL, join, and circuit machinery instead of a separate engine (\Cref{sec:epistemic}). The accepted fragment and its bounds are in \Cref{sec:limits}.
\end{itemize}

\textbf{Target workloads.} Avoiding repeated data-plane copies matters most where the symbolic workload is large and repeated: program-analysis queries over millions of ground atoms, probabilistic logic with many training iterations that share one compiled circuit, and knowledge-graph reasoning with trainable entity embeddings. Small tasks such as MNIST addition exercise the integration end-to-end but sit below this regime. They demonstrate correctness and the circuit-caching effect, with the measured transfer component secondary: approximately 10\% of a training step at a standard batch and lost in run-to-run variation when a step makes only a handful of neural$\to$symbolic handoffs (\Cref{sec:overhead}). Accordingly, xlog targets NVIDIA-GPU workloads whose byte working set and fixed capacities fit the selected route (\Cref{sec:limits}) and that need neural perception coupled to exact, verified probabilistic inference rather than fuzzy relaxations.

\Cref{sec:arch} presents the integration model and system architecture. \Cref{sec:lang} describes the xlog language. \Cref{sec:datalog} details the GPU-resident symbolic substrate. \Cref{sec:prob} presents verified knowledge compilation, the central technical result. \Cref{sec:nesy} ties the pieces together as end-to-end neurosymbolic learning. \Cref{sec:ecinduction} puts that surface on an external video-event benchmark, and \Cref{sec:maritime} carries the same induction protocol to a larger maritime corpus. \Cref{sec:epistemic} extends the substrate to epistemic and other reasoning modes. \Cref{sec:eval} reports evaluation results, \Cref{sec:related} surveys related work, \Cref{sec:limits} discusses limitations and future work, and \Cref{sec:conclusion} concludes.

\section{Integration Model and Architecture}\label{sec:arch}

\subsection{The integration model}

In xlog a neural network \emph{is} a predicate. A declaration binds a network to a predicate symbol; evaluating that predicate runs a forward pass, and the network's softmax over a label set can become a distribution over weighted facts. Typed device-buffer interfaces make those values available to the supported reasoning routes. Exact gradients of query values flow back through the probabilistic knowledge-compilation route; deterministic and epistemic execution reuse the integration surface without claiming that gradient semantics. Facts, circuit values, and gradient buffers remain device-resident at their data-plane handoffs, while orchestration and bounded control reads remain host responsibilities.
\subsection{GPU residency model}

xlog keeps runtime data-plane state in GPU memory and does not silently spill it or switch to a host executor. The provider owns one device handle, stream pool, asynchronous resource, runtime, and memory manager. Exceeding a configured byte budget raises \texttt{RESOURCE\_EXHAUSTED}; exhausting a fixed row, tuple, sample, or other count capacity raises \texttt{CAPACITY\_EXCEEDED}; and a bounded fixpoint that reaches its iteration limit raises \texttt{CONVERGENCE\_FAILURE}. Transfer counters delimit measured claims. In particular, the certified resident recursive and Monte Carlo sampled cores record zero tracked transfers, then return one bounded terminal receipt whose status is authoritative.

The payoff is bandwidth asymmetry. PCIe~4.0 $\times16$ delivers roughly 32\,GB/s, while GPU HBM provides 1--3\,TB/s. A single unnecessary D2H round-trip for a moderate relation (10M tuples at 16 bytes, ${\approx}160$\,MB) costs about 5\,ms. Keeping fact stores, deltas, circuit values, solver state, and gradient buffers on the GPU avoids copying those data planes through host memory; this claim does not erase the bounded control observations documented for each route.

\subsection{Crate architecture}

\begin{figure*}[tbp]
  \centering
  \begin{tikzpicture}[x=1mm,y=1mm,
      cell/.style={fnode,anchor=north west,font=\footnotesize,
                   inner xsep=2mm,inner ysep=1.6mm,minimum height=12mm},
      tag/.style={ftier,anchor=north west,text width=21mm,minimum height=12mm,
                  font=\footnotesize}]
    \node[tag] (t4tag) at (0,0)
      {\textbf{Tier 4}\\{\scriptsize\scshape interfaces}};
    \node[cell,text width=67.5mm] (t4c1) at (27,0)
      {Python API (PyO3)\\[0.6mm]\cratechip{pyxlog}};
    \node[cell,text width=67.5mm] (t4c2) at (100.5,0)
      {command-line interface\\[0.6mm]\cratechip{xlog-cli}};
    \node[tag] (t3tag) at (0,-19.5)
      {\textbf{Tier 3}\\{\scriptsize\scshape pipelines}};
    \node[cell,text width=43mm] (t3c1) at (27,-19.5)
      {deterministic\\[0.6mm]\cratechip{xlog-gpu}};
    \node[cell,text width=43mm] (t3c2) at (76,-19.5)
      {probabilistic $\cdot$ PIR $\cdot$ verified KC\\[0.6mm]\cratechip{xlog-prob}};
    \node[cell,text width=43mm] (t3c3) at (125,-19.5)
      {neural-symbolic\\[0.6mm]\cratechip{xlog-neural} \cratechip{xlog-induce}};
    \node[tag] (t2tag) at (0,-36.5)
      {\textbf{Tier 2}\\{\scriptsize\scshape subsystems}};
    \node[cell,text width=43mm] (t2c1) at (27,-36.5)
      {language frontend\\[0.6mm]\cratechip{xlog-logic}};
    \node[cell,text width=43mm] (t2c2) at (76,-36.5)
      {GPU runtime executor\\[0.6mm]\cratechip{xlog-runtime}};
    \node[cell,text width=43mm] (t2c3) at (125,-36.5)
      {GPU CDCL $\cdot$ local search\\[0.6mm]\cratechip{xlog-solve}};
    \node[tag] (t1tag) at (0,-53.5)
      {\textbf{Tier 1}\\{\scriptsize\scshape ir \& device\\[-0.4mm]\scriptsize\scshape services}};
    \node[cell,text width=43mm] (t1c1) at (27,-53.5)
      {relational / epistemic IRs\\[0.6mm]\cratechip{xlog-ir}};
    \node[cell,text width=43mm] (t1c2) at (76,-53.5)
      {CUDA wrapper, Arrow/DLPack\\[0.6mm]\cratechip{xlog-cuda}};
    \node[cell,text width=43mm] (t1c3) at (125,-53.5)
      {runtime statistics\\[0.6mm]\cratechip{xlog-stats}};
    \node[tag] (t0tag) at (0,-70.5)
      {\textbf{Tier 0}\\{\scriptsize\scshape core}};
    \node[cell,text width=141mm] (t0c1) at (27,-70.5)
      {scalar types $\cdot$ errors $\cdot$ symbol table $\cdot$ memory budgets\\[0.6mm]\cratechip{xlog-core}};
    \begin{scope}[on background layer]
      \node[fplane,fit=(t3tag)(t3c3)(t1tag)(t1c3),inner sep=2.2mm] (plane) {};
    \end{scope}
    \node[fplanetitle,anchor=east] at ([xshift=-4mm]plane.north east)
      {\planetitle{runtime data plane $\cdot$ gpu-resident}};
    \draw[farrow] (12.5,-12) -- (12.5,-19.5);
    \draw[farrow] (12.5,-31.5) -- (12.5,-36.5);
    \draw[farrow] (12.5,-48.5) -- (12.5,-53.5);
    \draw[farrow] (12.5,-65.5) -- (12.5,-70.5);
    \node[flabel,anchor=west] at (14,-14.2) {depends on};
    \node[flabel,anchor=south east] at (172,1.5)
      {host: orchestration $\cdot$ I/O $\cdot$ compilation};
    \node[flabel,anchor=north east] at (172,-84)
      {host-side foundation: shared types \& traits};
  \end{tikzpicture}
  \caption{xlog crate hierarchy. Each tier depends only on lower tiers; the
  shaded device plane marks GPU-resident runtime data planes, while the host
  provides orchestration, I/O, compilation, and bounded control reads.}
  \label{fig:arch}
\end{figure*}
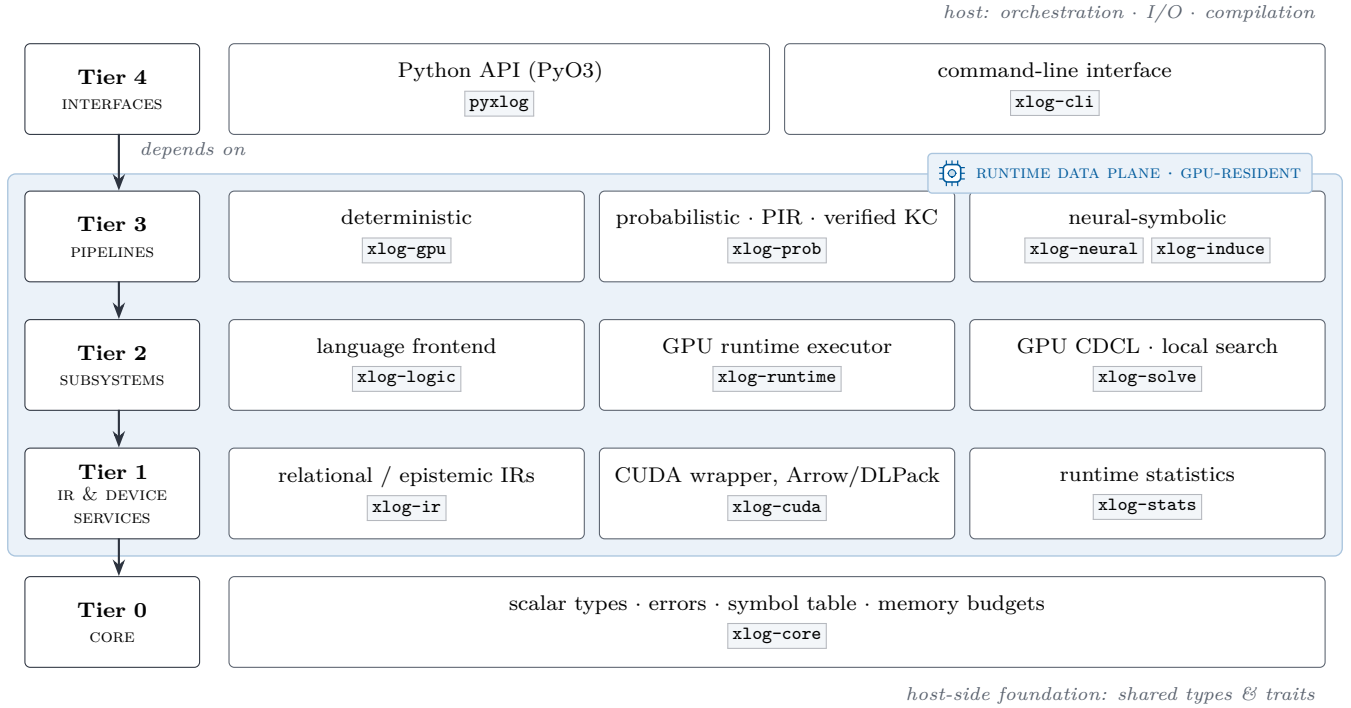

The workspace is organized into five tiers with strict layering: no crate depends on a peer or a higher tier (\Cref{fig:arch}). A leaf \emph{core} crate defines shared scalar types, error and memory-budget types, canonical boolean parsing, and a bidirectional symbol table for dictionary-encoded strings. Above it, a tier of intermediate representations and device services wraps CUDA via cudarc~\cite{cudarc}, resolves embedded kernel modules, and exposes Arrow and DLPack interoperability. Each CUDA provider owns its complete runtime resource graph; there is no process-wide runtime singleton or alternate provider path. The three core subsystems (the typed language frontend of \Cref{sec:lang}, the GPU runtime executor, and the GPU CDCL/local-search solver) compose into integrated pipelines for deterministic, probabilistic, and neural-symbolic execution, which a PyO3~\cite{pyo3} extension module (\texttt{pyxlog}) and a command-line binary surface to users.

\subsection{From program to GPU plan}

A program is compiled to a GPU-executable plan in five stages.

\textbf{Parsing.} A PEG parser built with pest produces an AST covering the full surface, including probabilistic facts, annotated disjunctions, epistemic literals, and neural-predicate declarations.

\textbf{Dependency analysis} builds a graph with positive, negative, aggregate, probabilistic, and modal edges and computes its strongly connected components (SCCs) with Tarjan's algorithm. It establishes a deterministic stratum order for the ordinary monotone core and classifies supported nonmonotone components for their dedicated plans.

\textbf{Lowering} fans out from the normalized AST into three reasoning IRs: deterministic and reduced ordinary rules become relational-IR trees, probabilistic provenance becomes PIR, and modal programs retain their semantics in EIR. Neural predicates compose with the applicable route. SAT/MaxSAT and verification use a separate shared service instead of a fourth reasoning IR.

\textbf{Optimization} pushes predicates through joins, while lowering chooses a deterministic greedy join order. Shape-specific selectivity rewrites cover recognized triangle and four-cycle plans; general dynamic-programming join ordering is not active. Eligible multiway rules are promoted to the WCOJ subsystem, and bound recursive queries can be rewritten via magic sets (\Cref{sec:datalog}).

\textbf{Plan assembly} emits the route-specific execution plan, whose relational work is dispatched to CUDA kernels. Supported negated recursion uses GPU-backed well-founded semantics instead of the ordinary semi-naive schedule.

\subsection{Reasoning IRs and circuit format}

Three reasoning IRs keep the backends decoupled; probabilistic knowledge compilation also emits a downstream circuit format. Each form has one role.

\textbf{RIR (Relational IR)} is the algebraic tree (\texttt{Scan}, \texttt{Filter}, \texttt{Project}, \texttt{Join}, \texttt{GroupBy}, \texttt{Union}, \texttt{Distinct}, \texttt{Diff}, \texttt{Fixpoint}, and the worst-case-optimal \texttt{MultiWayJoin}/\texttt{ChainJoin}) carrying cardinality, skew, and delta-vs-full metadata. Deterministic execution and reduced ordinary subprograms consume RIR; the other backends preserve additional semantics in their own IRs.

\textbf{PIR (Provenance IR)} is the weighted Boolean formula (\texttt{Lit}, \texttt{NegLit}, \texttt{And}, \texttt{Or}, \texttt{Decision}) derived from provenance tracking over a probabilistic program, capturing its structure without execution order.

\textbf{XGCF (xlog GPU Circuit Format)} is the compiled, device-resident form: a levelized DAG whose level offsets let every node at one topological level evaluate in a single kernel launch, with forward (log-space weighted model counting) and backward (adjoint) passes.

\textbf{EIR (Epistemic IR)} preserves modal operators for world-view reasoning and lowers to a dedicated GPU execution plan instead of ordinary relational algebra (\Cref{sec:epistemic}).

\textbf{GpuCnf and CDCL} form a shared solver service, not a fourth reasoning IR. Callers provide a device-resident CNF formula for SAT/MaxSAT search or verification. The CDCL service consumes that formula directly; XGCF remains the circuit format produced only by probabilistic knowledge compilation.

\section{The xlog Language}\label{sec:lang}

Probabilistic facts (\texttt{p::f.}) and neural predicate declarations (\texttt{nn/k}) are language-level constructs covered in \Cref{sec:prob,sec:nesy}.

\subsection{Design principles}

Three principles shape the surface, each chosen to enable GPU execution. \textbf{Typed predicates:} every predicate has a statically known signature over a closed set of scalar types (\texttt{u32}, \texttt{u64}, \texttt{i32}, \texttt{i64}, \texttt{f32}, \texttt{f64}, \texttt{bool}, \texttt{symbol}), supplied by a \texttt{pred} declaration or inferred during compilation, so argument mismatches are caught before any kernel is generated and allocation stays bounded and columnar. \textbf{Classified dependencies:} negation, aggregation, modal operators, and recursion interact through SCC analysis on the predicate dependency graph. The ordinary deterministic core requires a monotone or stratified schedule, while supported nonmonotone probabilistic and epistemic components enter dedicated plans, including GPU-backed well-founded semantics for supported recursion through negation. \textbf{One language for many paradigms:} probabilistic facts, annotated disjunctions, neural predicates, and SAT constraints share the syntactic core of deterministic Datalog; parsing, normalization, and dependency analysis are shared, while backend-specific lowering preserves the semantics needed by each plan. Unsupported shapes and unbounded terms are rejected before execution; there is no implicit fallback.

A minimal program exhibits the core surface:

\begin{lstlisting}[style=xlogstyle,
                 caption={Minimal xlog program: typed predicates, facts, a recursive rule, a query.},
                 label={lst:hello}]
// Typed predicates, facts, a rule, a query.
pred edge(u32, u32).
pred reach(u32, u32).

edge(1, 2). edge(2, 3). edge(3, 4).

reach(X, Y) :- edge(X, Y).
reach(X, Z) :- reach(X, Y), edge(Y, Z).

?- reach(1, N).
\end{lstlisting}

\subsection{Type system}\label{sec:lang:types}

Every predicate is assigned a schema over the eight scalar types, with \texttt{f32}/\texttt{f64} following IEEE~754. A \texttt{pred} declaration supplies that schema explicitly; when it is omitted, compilation infers the schema from facts, rule heads, typed body columns, and arithmetic bindings. String literals (\texttt{"alice"}) map to \texttt{symbol} through a global symbol table, giving the runtime dense integer identifiers while preserving source-level readability; \texttt{domain} declarations introduce named aliases. Type consistency is enforced by predicate schemas and lowering-time checks: variable occurrences must agree with known column types, constants are validated against their destination columns, and arithmetic expressions preserve scalar types.

\begin{lstlisting}[style=xlogstyle,
                 caption={A well-typed xlog program: typed bridge predicate.},
                 label={lst:types-ok}]
pred node(u32, symbol).
pred connected(u32, u32).
pred bridge(u32, u32).

node(1, "alice"). node(2, "bob").
connected(1, 2).

bridge(A, B) :-
  connected(A, B),
  node(A, _), node(B, _).
\end{lstlisting}

Declaring \texttt{bridge} so that its head argument is constrained at two incompatible types (e.g.\ \texttt{symbol} in the head but \texttt{u32} through \texttt{connected}) rejects the program at compile time.

\subsection{Expressions, arithmetic, and user-defined functions}\label{sec:lang:arith}

Rule bodies may contain arithmetic, comparisons, and calls. Arithmetic bindings use the \texttt{is} operator (\texttt{D is X + Y}); the right-hand side supports \texttt{+ - * / \%}, the built-ins \texttt{abs}, \texttt{min}, \texttt{max}, \texttt{pow}, \texttt{cast}, and user-defined functions. Comparisons compile to \texttt{Filter} nodes and push below joins where profitable. Float comparison uses a total ordering, so \texttt{NaN} and signed zeros yield deterministic results instead of contaminating downstream aggregations.

\begin{lstlisting}[style=xlogstyle,
                 caption={Arithmetic in rule bodies: pairwise proximity via squared Euclidean distance.},
                 label={lst:arith}]
pred point(u32, f64, f64).
pred close(u32, u32).

point(1, 0.0, 0.0).
point(2, 0.5, 0.5).
point(3, 5.0, 5.0).

close(A, B) :-
  point(A, Xa, Ya),
  point(B, Xb, Yb),
  A < B,
  D is (Xa - Xb) * (Xa - Xb)
     + (Ya - Yb) * (Ya - Yb),
  D < 1.0.
\end{lstlisting}

A user-defined function (UDF) is introduced with \texttt{func}, with optional parameter types and a return type after \texttt{->}. Arithmetic bodies (optionally with \texttt{if/then/else}) inline into the same \texttt{Expr} trees as built-in arithmetic and participate in the same predicate-pushdown and expression-level optimization. A predicate body, as in \texttt{func get\_parent(Child) = Parent :- parent(Child, Parent).}, contributes its relational literals immediately before the calling \texttt{is} binding in an ordinary rule or integrity constraint. Its relation may contribute zero, one, or many caller rows; the function syntax does not impose uniqueness, while ordinary set semantics deduplicates identical projected tuples. Each invocation alpha-renames its non-parameter result and body-local variables, and multiple calls expand from left to right. Nested expansion is depth-bounded; relational calls in conditional result branches and non-term arithmetic arguments are rejected rather than hoisted or coerced with changed semantics.

\begin{lstlisting}[style=xlogstyle,
                 caption={User-defined function: \texttt{clamp01} normalizes a raw score into the unit interval.},
                 label={lst:udf}]
pred raw_score(u32, f64).
pred norm_score(u32, f64).

func clamp01(X: f64) -> f64 =
  if X < 0.0 then 0.0
  else if X > 1.0 then 1.0
  else X.

raw_score(1, -0.3).
raw_score(2, 0.7).
raw_score(3, 1.4).

norm_score(Id, N) :-
  raw_score(Id, R), N is clamp01(R).
\end{lstlisting}

\subsection{Modules and imports}\label{sec:lang:modules}

Programs decompose into \texttt{.xlog} modules on a search path. The \texttt{use} directive loads a module (\texttt{use graph.}) or selectively imports names (\texttt{use utils/math::\{abs, clamp\}.}); a \texttt{private} modifier hides declarations from importers. The resolver traverses the import graph, detects cycles, validates participating declarations, and performs bounded cross-module schema inference for undeclared predicate contributions before merging them. The resulting logical program then undergoes full compiler type inference, dependency analysis, and route-specific execution planning, so imported rules participate as if written inline.

\begin{lstlisting}[style=xlogstyle,
                 caption={Module \texttt{graph\_lib.xlog}: predicate declarations and recursive reach rules.},
                 label={lst:graph-lib}]
// library: graph_lib.xlog
pred edge(u32, u32).
pred reach(u32, u32).

edge(1, 2). edge(2, 3). edge(3, 4).

reach(X, Y) :- edge(X, Y).
reach(X, Z) :- reach(X, Y), edge(Y, Z).
\end{lstlisting}

\begin{lstlisting}[style=xlogstyle,
                 caption={Entry \texttt{graph\_main.xlog}: imports \texttt{graph\_lib} and queries \texttt{reach}.},
                 label={lst:graph-main}]
// entry: graph_main.xlog
use graph_lib.

?- reach(1, N).
\end{lstlisting}

\subsection{Aggregations and constraints}\label{sec:lang:agg}

Aggregation operators---\texttt{count}, \texttt{sum}, \texttt{min}, \texttt{max}, \texttt{logsumexp}---appear at rule heads with an explicit aggregation variable and lower to \texttt{GroupBy} nodes executed as radix-sort-and-reduce kernels; the stratifier rejects aggregation inside a recursive SCC. Integrity constraints are headless rules (\texttt{:- body.}) asserting that \texttt{body} is unsatisfiable once evaluation reaches fixpoint; a violation raises a diagnostic. \texttt{logsumexp} is included because it is the workhorse aggregation for probabilistic circuits, where log-space numerical stability is essential (\Cref{sec:prob}).

\begin{lstlisting}[style=xlogstyle,
                 caption={Stratified aggregation: per-source out-degree with an integrity constraint.},
                 label={lst:agg}]
pred edge(u32, u32).
pred degree(u32, u32).

edge(1, 2). edge(1, 3).
edge(1, 4). edge(2, 3).

degree(X, count(Y)) :- edge(X, Y).

:- edge(X, _), not degree(X, _).

?- degree(X, N).
\end{lstlisting}

\subsection{Completeness over finite domains}\label{sec:lang:completeness}

The surface extends toward language completeness while preserving GPU-native execution: accepted constructs normalize into the typed AST and lower through their semantic backend, while forms that cannot be lowered safely are rejected at compile time, since there is no hidden CPU interpreter to execute them. The accepted extensions are finite typed lists (\texttt{[]}, \texttt{[H|T]}, the \texttt{list<T>} column type), statically-resolvable meta-predicates (\texttt{ground}, \texttt{functor}, \texttt{=..}, source-fact \texttt{findall}, \texttt{maplist}), explicit stratified negation-as-failure over deterministic relations, magic-set rewriting of bound recursive queries, and finite probabilistic aggregates. Deterministic extensions lower to relational joins and aggregations in RIR; probabilistic aggregates preserve their probabilistic intermediate representation. Open-ended generators, dynamic database mutation, unrestricted \texttt{call/N}, and runtime-variable predicate names remain outside the contract.

\section{GPU-Native Symbolic Substrate}\label{sec:datalog}

The deterministic Datalog engine supplies CUDA relational operators and device-resident relation storage used across the system. The algorithmic approach---semi-naive fixpoint iteration over stratified programs---is well established; the contribution here is engineering. The ordinary executor is host-orchestrated. For eligible recursive query plans, automatic selection instead uses a certified resident conditional graph: its core runs one recorded launch and synchronization with zero tracked host--device transfers, then returns one pinned terminal receipt carrying status, counters, trace metadata, and schema selection rather than tuple data.

\subsection{Semi-naive evaluation on GPU}

The executor processes an execution plan as strata ordered by the dependency analysis of \Cref{sec:arch}. Non-recursive SCCs evaluate each rule once, dispatching its RIR tree to the appropriate kernel and merging results per head predicate. Recursive SCCs run semi-naive iteration (\Cref{alg:seminaive}): each rule is re-evaluated through delta-rewritten variants (one per recursive scan occurrence, so a self-join such as \texttt{p(X,Y), p(Y,Z)} contributes two variants) and the new tuples are unioned, differenced against the full relation, and deduplicated, all on-device, until no predicate produces new tuples. The profiler records per-operator timing, row counts, and peak memory, feeding the optimizer.

\begin{algorithm}[htbp]
\caption{GPU semi-naive evaluation of a recursive SCC. All buffers are device-resident.}
\label{alg:seminaive}
\begin{algorithmic}[1]
\Require SCC rules $\mathcal{R}$; relation store $S$ (device-resident)
\Ensure least fixpoint of $\mathcal{R}$ merged into $S$
\For{each rule $r \in \mathcal{R}$}
  \State $\Delta_r \gets \textsc{Eval}(r, S) \setminus S$ \Comment{seed $\Delta$; set difference on device}
\EndFor
\Repeat
  \For{each rule $r \in \mathcal{R}$, recursive scan $i$ in $r$}
    \State $T_{r,i} \gets \textsc{Eval}(r[\Delta \text{ at } i], S)$ \Comment{one $\Delta$-variant per self-join}
  \EndFor
  \For{each head predicate $p$}
    \State $\Delta_p \gets \textsc{Dedup}\big(\big(\textstyle\bigcup_{r,i \text{ for } p} T_{r,i}\big) \setminus S_p\big)$
    \State $S_p \gets S_p \cup \Delta_p$ \Comment{union + sorted dedup on device}
  \EndFor
\Until{all $\Delta_p = \emptyset$ \textbf{or} iteration cap reached}
\State free all $\Delta$ buffers
\end{algorithmic}
\end{algorithm}

\subsection{Relational kernels}

xlog implements its relational algebra as direct CUDA kernels rather than wrappers over cuBLAS, cuSPARSE, or Thrust. Bespoke kernels make allocation, synchronization, and host staging explicit and keep relational intermediate buffers on-device. \Cref{tab:kernels} summarizes them.

\begin{table}[htbp]
  \centering
  \caption{Core relational CUDA kernels.}
  \label{tab:kernels}
  \small
  \begin{tabularx}{\linewidth}{@{}l L@{}}
    \toprule
    \textbf{Kernel} & \textbf{Purpose} \\
    \midrule
    \texttt{join}    & Hash join with linked-list collision chains and FNV-1a composite hashing over raw key bytes, so all scalar types share one path; count\,$\to$\,scan\,$\to$\,materialize variants for inner and left-outer joins. \\
    \texttt{sort}    & Stable 4-bit radix sort (8 passes over 32-bit keys): per-pass histogram, prefix sum, scatter. \\
    \texttt{filter}  & Comparison operators for column-vs-constant and column-vs-column, with stream compaction via prefix sum. \\
    \texttt{groupby} & Sorted-input grouped aggregation: boundary detection, then per-group Count/Sum/Min/Max/LogSumExp reduction. \\
    \texttt{dedup}   & Sort-based deduplication and set difference with type-aware equality, including IEEE~754 float handling. \\
    \texttt{set\_ops} & Union (concat\,+\,sort\,+\,dedup) and set difference via sorted-array binary search. \\
    \texttt{scan}    & Blelloch parallel prefix sum, the building block for filter compaction, dedup, and group-by. \\
    \texttt{pack}    & Multi-column key packing into row-major byte arrays with FNV-1a hashing. \\
    \bottomrule
  \end{tabularx}
\end{table}

The filter kernel makes a correctness-critical choice for floating point. Equality uses standard IEEE~754 semantics ($\mathrm{NaN}\neq\mathrm{NaN}$); ordered comparisons use a total-ordering transform that maps an \texttt{f64} bit pattern to an \texttt{i64} whose integer order matches the IEEE~754 \texttt{totalOrder} predicate ($-\mathrm{NaN} < -\mathrm{Inf} < $ negatives $< -0.0 < +0.0 <$ positives $< +\mathrm{Inf} < +\mathrm{NaN}$), so Datalog programs produce deterministic results for all floating-point inputs.

\subsection{Adaptive join planning}

The optimizer performs predicate pushdown using runtime statistics, decomposing filters above joins into left-, right-, and cross-predicates. Lowering chooses join order with a deterministic greedy heuristic. A selectivity pass recognizes triangle and four-cycle shapes, and prior execution statistics can be exported and merged before a later compilation. A general dynamic-programming join ordering is not active; configuration fields that resemble such a threshold do not change this planning path.

\subsection{Worst-case-optimal joins}

Binary-join plans can materialize intermediates asymptotically larger than the final result, the classic pathology for cyclic queries such as triangle enumeration. This is not a corner case for xlog's target workloads: program-analysis queries (points-to, call-graph and reachability inference over code-dependency graphs) can contain cyclic, skewed joins where intermediate size governs cost. xlog addresses this with a worst-case-optimal join (WCOJ) subsystem~\cite{ngo2018wcoj,veldhuizen2014leapfrog,wang2023freejoin}. A pure-Rust hypergraph planner analyzes rule bodies for multiway-join eligibility, derives a deterministic variable ordering from cardinality and selectivity statistics, and emits a \texttt{MultiWayJoin}/\texttt{ChainJoin} node when promotion preserves output semantics; other rules retain their ordinary binary-join plan.

The GPU kernel executes the join via count\,$\to$\,device prefix scan\,$\to$\,materialize over flat sorted-column storage, with a four-byte metadata D2H total and no count-vector transfer. Coverage spans the triangle, 4-cycles, $K$-clique templates ($K{=}5$--$8$), recursive/SCC integration, and helper-relation splitting that exposes inner skew to root-level histograms. A default-on skew classifier decides per query whether to dispatch WCOJ: high-skew inputs take the WCOJ path, while uniform or empty inputs select the binary-join chain. \Cref{sec:wcoj-eval} reports the measured speedup.

\subsection{Runtime optimization}

Interactive and long-running deployments (REPL sessions, iterative solver loops, incremental training) repeatedly evaluate similar queries over relations that change little between calls. A stateless executor rebuilds join indices and recomputes shared subplans each time, paying the dominant cost repeatedly. Three optimizations exploit this between-call stability, each with explicit controls and zero added data-plane transfers. Common-subexpression elimination caches safe deterministic subplans within an evaluation. Adaptive re-optimization adopts a compiler-supplied candidate plan when mis-plan telemetry crosses a configurable ratio, with output-equivalence checks and rollback. And a persistent hash-index manager, keyed on relation generation, schema, and device, retains built indices across repeated sessions with deterministic LRU eviction (\Cref{sec:eval}).

\subsection{Reversible symbols}

Datalog programs operate on strings, but GPU kernels operate on fixed-width numeric columns. xlog interns each unique string to a dense, sequential \texttt{u32} identifier and resolves it back in $O(1)$ at output time. Symbol columns then \emph{are} ordinary \texttt{u32} columns (join, sort, filter, and dedup kernels operate on them with no special-casing) and the string table stays host-side, since variable-length data is needed only at ingestion and output. The \texttt{symbol} type maps to Arrow's \texttt{Dictionary(UInt32, Utf8)}, so export to columnar consumers (cuDF, Polars, DuckDB) is zero-copy while preserving the compact internal representation.

\section{Verified Knowledge Compilation}\label{sec:prob}

Knowledge compilation is the route that makes a neural network and a probabilistic logic program differentiable end-to-end. A network's softmax becomes a distribution over weighted facts; provenance tracking yields a weighted Boolean formula; and that formula compiles into an arithmetic circuit whose forward pass is exact weighted model counting and whose backward pass yields exact gradients. Probabilistic systems such as ProbLog~\cite{problog2} follow this compile-once/evaluate-many model on the CPU. xlog instead keeps the compilation and evaluation data planes in CUDA buffers while the host orchestrates stages and reads bounded control state. Before use, xlog certifies the final smoothed circuit against its back-end source formula; this certificate does not cover provenance extraction or CNF encoding.

\subsection{The compilation pipeline}

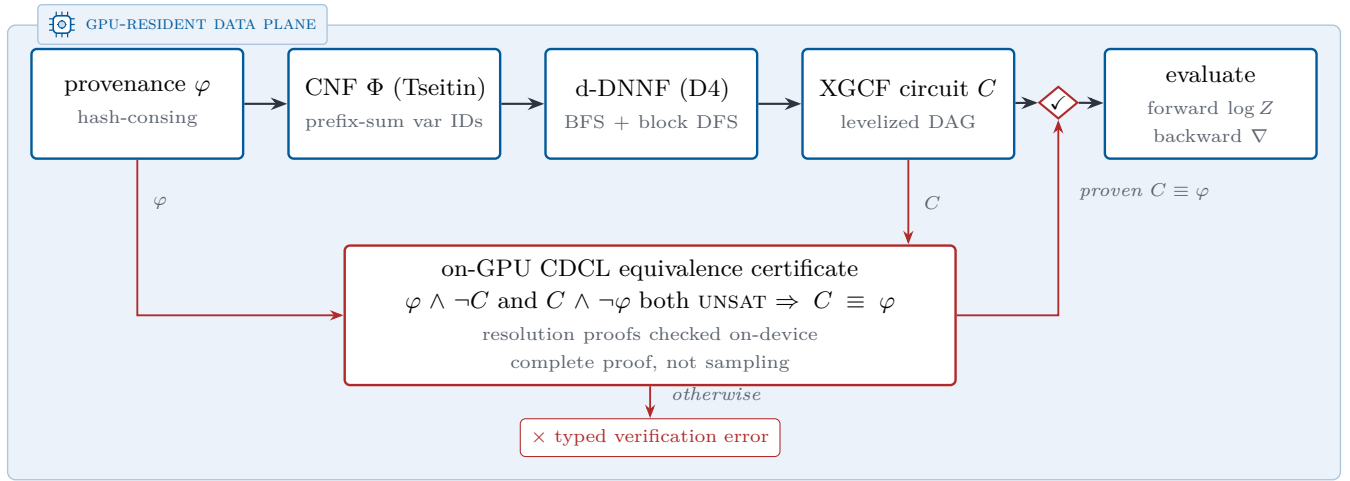
\begin{figure*}[tbp]
  \centering
  \begin{tikzpicture}[x=1mm,y=1mm,
      stage/.style={fproc,anchor=north west,text width=24mm,
                    inner xsep=2mm,inner ysep=1.8mm,minimum height=14.5mm}]
    \node[stage] (s1) at (0,0)
      {provenance $\varphi$\\[0.4mm]\fsubt{hash-consing}};
    \node[stage] (s2) at (34,0)
      {CNF $\Phi$ (Tseitin)\\[0.4mm]\fsubt{prefix-sum var IDs}};
    \node[stage] (s3) at (68,0)
      {d-DNNF (D4)\\[0.4mm]\fsubt{BFS + block DFS}};
    \node[stage] (s4) at (102,0)
      {XGCF circuit $C$\\[0.4mm]\fsubt{levelized DAG}};
    \node[stage] (s5) at (142,0)
      {evaluate\\[0.4mm]\fsubt{forward $\log Z$}\\[-0.2mm]\fsubt{backward $\nabla$}};
    \node[diamond,aspect=1.2,draw=figaccent,fill=white,line width=0.8pt,
          inner sep=0.5pt] (gate) at (136,-7.25) {\scriptsize\checkmark};
    \draw[farrow] (s1) -- (s2);
    \draw[farrow] (s2) -- (s3);
    \draw[farrow] (s3) -- (s4);
    \draw[farrow] (s4) -- (gate);
    \draw[farrow] (gate) -- (s5);
    \node[fcert,anchor=north,text width=78mm,inner ysep=1.8mm] (cert) at (82,-26)
      {on-GPU CDCL equivalence certificate\\[0.4mm]
       $\varphi\wedge\lnot C$ and $C\wedge\lnot\varphi$ both \textsc{unsat}
       $\Rightarrow C\equiv\varphi$\\[0.4mm]
       \fsubt{resolution proofs checked on-device}\\[-0.2mm]
       \fsubt{complete proof, not sampling}};
    \draw[fver] (s1.south) |- (cert.west);
    \node[flabel,anchor=west] at (15,-20.5) {$\varphi$};
    \draw[fver] (s4.south) -- (s4.south |- cert.north);
    \node[flabel,anchor=west] at (117,-20.5) {$C$};
    \draw[fver] (cert.east) -| (gate.south);
    \node[flabel,anchor=west] at (137.5,-19) {proven $C\equiv\varphi$};
    \node[fnode,anchor=north,draw=figaccent,font=\scriptsize,
          text=figaccent,inner ysep=1.4mm,minimum height=5mm] (typederr) at (82,-49)
      {$\times$ typed verification error};
    \draw[fver] (cert.south) -- (typederr.north);
    \node[flabel,anchor=west] at (83.5,-45.8) {otherwise};
    \begin{scope}[on background layer]
      \node[fplane,fit=(s1)(s5)(cert)(typederr),inner sep=3mm] (plane) {};
    \end{scope}
    \node[fplanetitle,anchor=west] at ([xshift=4mm]plane.north west)
      {\planetitle{gpu-resident data plane}};
  \end{tikzpicture}
  \caption{Verified knowledge-compilation pipeline. The data plane uses CUDA
  kernels and device-resident state under host orchestration. The final smoothed
  circuit $C$ reaches evaluation only through the equivalence gate (\checkmark):
  an on-GPU CDCL solver proves $\varphi\wedge\lnot C$ and
  $C\wedge\lnot\varphi$ both unsatisfiable; the host observes bounded status and
  error scalars, and any failed or incomplete proof returns a typed error.}
  \label{fig:kc}
\end{figure*}

The pipeline transforms a probabilistic program into a GPU-resident arithmetic circuit (the XGCF format) through five host-orchestrated stages (\Cref{fig:kc}). Their intermediate data stays in device buffers, while solver completion is observed through bounded status and error scalars.

\textbf{Provenance extraction.} Provenance over deterministic evaluation produces a weighted Boolean formula~\cite{green2007provenance}. An on-device interner hash-conses node batches through an atomic GPU hash table, deduplicating structurally identical subformulas; this is essential because grounding can produce exponentially many redundant subformulas, and interning on the host would force the uninterned graph into host memory---reintroducing the very transfer the pipeline exists to avoid.

\textbf{CNF encoding.} A Tseitin~\cite{tseitin} transformation lowers the formula to CNF on the GPU, using parallel prefix sums to assign compact, gap-free variable IDs and emit clauses into a device-resident compressed-sparse-row structure.

\textbf{D4 compilation.} The CNF compiles into a Decision-DNNF (d-DNNF) circuit~\cite{darwiche2001dnnf} via a GPU-native variant of the D4 algorithm~\cite{d4compiler}: a BFS frontier exposes parallelism, per-block DFS workers perform component detection, decision-variable selection by VSIDS-like activity~\cite{chaff}, and recursive decomposition. A smoothing pass forces every branch of an OR/decision node to mention the same variable set, and the result is levelized so each topological level evaluates in one kernel launch. Decomposability, determinism, and smoothness (the structural properties that make weighted model counting over a Decision-DNNF correct~\cite{darwicheMarquis2002,chaviraDarwiche2008}) are established by construction. Certification runs after smoothing and therefore gates the exact final circuit that is cached and evaluated.

\subsection{Verification: a machine-checked certificate}

xlog does not trust the final compilation stage; it proves the final smoothed circuit $C$ equivalent to its back-end source formula $\varphi$ (\Cref{alg:verifiedkc}). It solves $\varphi \land \lnot C$ and $C \land \lnot \varphi$ on a GPU CDCL solver, and both must be unsatisfiable for $C \equiv \varphi$. UNSAT results carry resolution proofs checked on-device. The host reads bounded status and error scalars before accepting that GPU proof result; a wrong, incomplete, or budget-limited result returns a typed error and never reaches circuit caching or evaluation.

The certificate says that the circuit computes the same Boolean function as $\varphi$; the count-correctness invariants are guaranteed separately, by construction. The compilation back-end thus becomes a machine-checked stage instead of an assumption, closing the failure mode in which a silently miscompiled circuit would corrupt every downstream probability and gradient. The check covers the back-end stage specifically: provenance extraction and CNF encoding are trusted, and the certificate says nothing about them. CPU-side certified knowledge compilation exists~\cite{capelli2019}; the contribution here is performing the check on-device, sharing the GPU substrate. The same device-resident CDCL substrate is reused for SAT/MaxSAT queries and for epistemic candidate validation (\Cref{sec:epistemic}).

\begin{algorithm}[htbp]
\caption{Verified knowledge compilation. The final smoothed circuit is accepted only if proven equivalent to its back-end source formula.}
\label{alg:verifiedkc}
\begin{algorithmic}[1]
\Require weighted Boolean formula $\varphi$ (device-resident)
\Ensure verified circuit $C$ (XGCF), or typed error
\State $\varphi \gets \textsc{InternHashCons}(\varphi)$ \Comment{dedup via atomic GPU hash table}
\State $\Phi \gets \textsc{TseitinCNF}(\varphi)$ \Comment{prefix-sum variable IDs; CSR clauses}
\State $C \gets \textsc{D4}(\Phi)$; $C \gets \textsc{SmoothLevelize}(C)$
\State $u_1 \gets \textsc{CDCL}(\varphi \land \lnot C)$;\quad $u_2 \gets \textsc{CDCL}(C \land \lnot \varphi)$
\State observe bounded status and error scalars for $u_1,u_2$ on host
\If{$u_1 \neq \textsc{Unsat}$ \textbf{or} $u_2 \neq \textsc{Unsat}$}
  \State \Return typed verification error
\EndIf
\State check resolution proofs of $u_1, u_2$ on-device
\State \Return $C$ \Comment{cache the exact circuit that was certified}
\end{algorithmic}
\end{algorithm}

The verified circuit supports a level-parallel forward pass (log-space weighted model counting, one kernel per level, yielding $\log Z$) and a reverse-order backward pass that accumulates per-variable gradients. Both leave their results in GPU-resident buffers consumable directly by PyTorch via DLPack (\Cref{sec:nesy}).

\subsection{Circuit caching}

Building and certifying a circuit is the most expensive stage, on the order of a minute for a cold start on the MNIST-addition benchmark, and a naive implementation pays it every epoch. Measured separately, it is the equivalence proof and not the D4 compile that dominates that time (\Cref{sec:overhead}). The key observation is that the circuit \emph{structure} depends on the propositional structure of the grounded program and the evidence pattern, but \emph{not} on the numerical weights. When neural parameters change between iterations, the weights on probabilistic facts change while the Boolean formula and its compiled circuit stay identical. The certified final smoothed circuit is therefore cached, keyed by a structural hash of the provenance graph and evidence configuration; subsequent iterations update only the weight buffers and run the forward/backward kernels. On MNIST addition that amortization is what the circuit-cache ablation of \Cref{sec:cache-eval} measures.

\subsection{Monte Carlo alternative}

When the user selects approximate inference, xlog provides a GPU-parallel Monte Carlo engine for its supported fragment. It draws values for probabilistic facts on the GPU and evaluates the deterministic program in each sampled world, using rejection sampling or, when every evidence atom is a Bernoulli fact, evidence clamping that forces observed values and eliminates rejection waste. Both methods report confidence intervals. The resident sampler evaluates all worlds in a single device launch with a bounded device-side fixpoint. Its sampled core records zero tracked transfers; after synchronization, one pinned receipt returns status, counts, and trace metadata. That receipt's convergence status is authoritative: reaching the iteration bound returns \texttt{CONVERGENCE\_FAILURE}, even if partial counters exist. Fixed-count exhaustion returns \texttt{CAPACITY\_EXCEEDED}, byte-budget exhaustion returns \texttt{RESOURCE\_EXHAUSTED}, and unsupported fragments return their typed diagnostic; none selects a host execution path.

\section{Neurosymbolic Integration}\label{sec:nesy}

This section is where the relational substrate (\Cref{sec:datalog}), verified probabilistic knowledge compilation (\Cref{sec:prob}), and device-buffer interoperability meet. Network outputs feed weighted facts into the probabilistic route; the final smoothed circuit is certified before use; and DLPack carries device gradient buffers into the neural optimizer without copying them through host memory. Host orchestration and bounded control reads remain outside that zero-copy data-plane handoff.

\subsection{Neural predicates}\label{subsec:neural-pred}

In xlog a neural network \emph{is} a predicate. The \texttt{nn/4} declaration embeds a network directly into the program:

\begin{lstlisting}[style=xlogstyle,
                   caption={Neural predicate declaration (\texttt{nn/4}).},
                   label={lst:nn-decl}]
nn(network_name, [input_vars], output_var,
   [output_labels]) :: predicate(args).
\end{lstlisting}

This is not a foreign-function escape hatch: it states that evaluating \texttt{predicate(args)} requires a forward pass through the named network, whose softmax over the label set becomes a distribution over probabilistic facts feeding the knowledge-compilation pipeline of \Cref{sec:prob}. On the Python side a PyTorch module is registered against the predicate:

\begin{lstlisting}[style=pythonstyle,
                   caption={Registering a PyTorch module against a neural predicate.},
                   label={lst:nn-register}]
program.register_network("mnist_net", net, optimizer)
\end{lstlisting}

Registration goes by type rather than by name. A caller may state the arity of the network's predicate, and that claim is checked against every \texttt{nn/4} declaration bound to the network in the program and rejected on mismatch: the program's own declaration is the authority. Per-argument catalog sort identifiers may accompany the arity (one per declared argument, refused without an arity or at the wrong length) and an opaque content hash may identify the registered network, so a retrained one mints a new identity when it is registered again. The engine interprets neither; both are carried for consumers, which read them back through a queryable metadata view together with each declaration's predicate, predicate arity, input arity, and label set. A downstream rule learner can therefore check a candidate against what the program declares instead of against a naming convention---the property \Cref{subsec:engine-mode} relies on.

The dataflow runs in two directions. \textbf{Forward:} when evaluation reaches an \texttt{nn/4}-backed predicate, the registered module runs, and its softmax vector is decomposed into weighted probabilistic facts, one per label, and fed into the knowledge-compilation pipeline of \Cref{sec:prob}, whose circuit cache carries across iterations. \textbf{Backward:} after the circuit computes the forward log-probability and backward gradients via level-parallel kernels, the per-leaf gradient buffers are exported as DLPack tensors and injected into PyTorch's autograd graph, and the optimizer updates the network as usual. The GIL is released during GPU work, so CUDA execution does not block the interpreter. A neural predicate is not confined to a query head: it may also appear inside rule bodies, so that the trainable object becomes a whole rule and not only a single perceptual predicate (\Cref{subsec:trainable-bodies}).

\subsection{End-to-end training}

The canonical example is MNIST addition (the DeepProbLog benchmark): given two digit images, predict their sum, with supervision only on sums: the network never sees individual digit labels. The entire reasoning structure is two rules:

\begin{lstlisting}[style=xlogstyle,
                   caption={MNIST addition: a CNN-backed digit predicate and an addition rule.},
                   label={lst:mnist-prog}]
nn(mnist_net, [X], Y, [0,1,2,3,4,5,6,7,8,9])
    :: digit(X, Y).
addition(X, Y, Z) :-
    digit(X, D1), digit(Y, D2), Z is D1 + D2.
\end{lstlisting}

The \texttt{nn/4} declaration connects the CNN to \texttt{digit/2}; the addition rule computes every consistent decomposition of a sum through probabilistic marginalization. Each query \texttt{addition(i, j, s)} trains by minimizing $-\log P(\texttt{addition}(i, j, s))$ under the compiled circuit. The driver compiles the program, registers the network, and runs the loop, which batches queries sharing a circuit template and runs forward/backward over the cached circuit:

\begin{lstlisting}[style=pythonstyle,
                   caption={Python driver: compile, register, train.},
                   label={lst:mnist-train}]
program = pyxlog.Program.compile(SRC)
program.register_network("mnist_net", net, optimizer)
program.add_tensor_source("train", train_images)
history = pyxlog.train_model_tensor(
    program, queries,
    epochs=epochs, batch_size=batch_size)
\end{lstlisting}

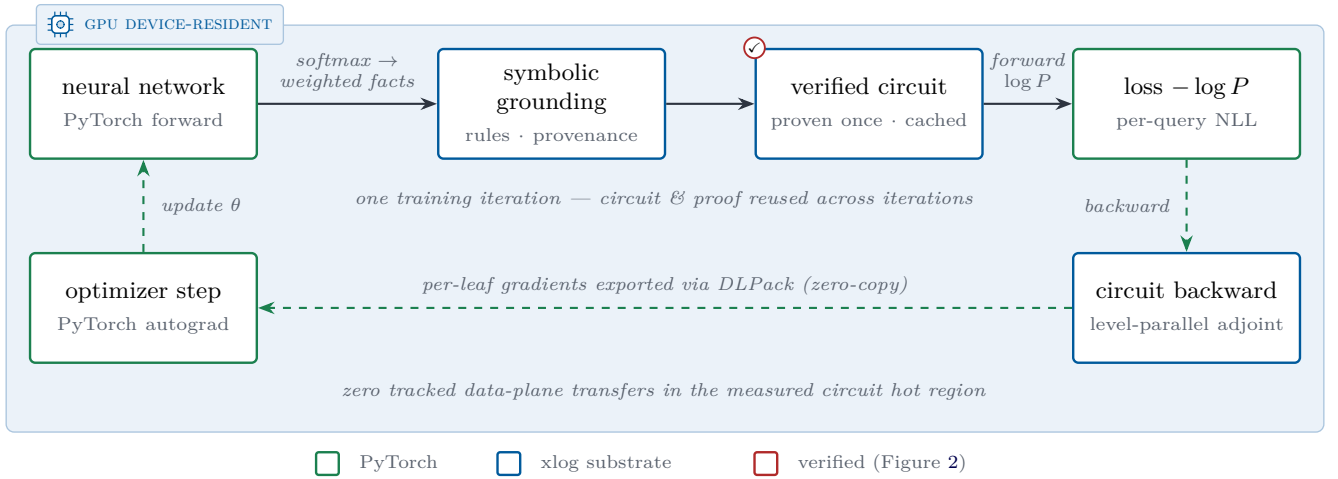
\begin{figure*}[tbp]
  \centering
  \begin{tikzpicture}[x=1mm,y=1mm,
      ring/.style={anchor=north west,text width=26mm,
                   inner xsep=2mm,inner ysep=1.8mm,minimum height=14.5mm}]
    \node[fneural,ring] (net) at (0,0)
      {neural network\\[0.4mm]\fsubt{PyTorch forward}};
    \node[fproc,ring] (ground) at (54,0)
      {symbolic grounding\\[0.4mm]\fsubt{rules $\cdot$ provenance}};
    \node[fproc,ring] (circ) at (96,0)
      {verified circuit\\[0.4mm]\fsubt{proven once $\cdot$ cached}};
    \node[fneural,ring] (loss) at (138,0)
      {loss $-\log P$\\[0.4mm]\fsubt{per-query NLL}};
    \node[circle,draw=figaccent,fill=white,line width=0.7pt,inner sep=0.4pt,
          font=\tiny] at (circ.north west) {\checkmark};
    \node[fproc,ring] (back) at (138,-27)
      {circuit backward\\[0.4mm]\fsubt{level-parallel adjoint}};
    \node[fneural,ring] (opt) at (0,-27)
      {optimizer step\\[0.4mm]\fsubt{PyTorch autograd}};
    \draw[farrow] (net) -- node[flabel,above=0.4mm,text width=21mm]
      {softmax $\to$ weighted facts} (ground);
    \draw[farrow] (ground) -- (circ);
    \draw[farrow] (circ) -- node[flabel,above=0.4mm,text width=11mm]
      {forward $\log P$} (loss);
    \draw[fback] (loss) -- node[flabel,anchor=east,xshift=-1mm] {backward} (back);
    \draw[fback] (back) -- node[flabel,above=0.4mm]
      {per-leaf gradients exported via DLPack (zero-copy)} (opt);
    \draw[fback] (opt) -- node[flabel,anchor=west,xshift=1mm]
      {update $\theta$} (net);
    \node[flabel] at (84,-20)
      {one training iteration --- circuit \& proof reused across iterations};
    \node[flabel] (foot) at (84,-45.5)
      {zero tracked data-plane transfers in the measured circuit hot region};
    \begin{scope}[on background layer]
      \node[fplane,fit=(net)(loss)(back)(opt)(foot),inner sep=3mm] (plane) {};
    \end{scope}
    \node[fplanetitle,anchor=west] at ([xshift=4mm]plane.north west)
      {\planetitle{gpu device-resident}};
    \draw[figgreen,line width=0.9pt,fill=white,rounded corners=1pt]
      (38,-56.5) rectangle +(3,3);
    \node[fsub,anchor=west] at (42.5,-55) {PyTorch};
    \draw[figblue,line width=0.9pt,fill=white,rounded corners=1pt]
      (62,-56.5) rectangle +(3,3);
    \node[fsub,anchor=west] at (66.5,-55) {xlog substrate};
    \draw[figaccent,line width=0.9pt,fill=white,rounded corners=1pt]
      (96,-56.5) rectangle +(3,3);
    \node[fsub,anchor=west] at (100.5,-55) {verified (\Cref{fig:kc})};
  \end{tikzpicture}
  \caption{One neurosymbolic training iteration. Neural tensors, circuit values,
  and gradient buffers occupy one CUDA device (shaded plane); the measured circuit
  hot region records zero tracked data-plane transfers. The host still orchestrates
  stages and observes bounded control state. The final smoothed circuit is compiled
  and certified once (\checkmark, \Cref{fig:kc}), then reused across iterations, and
  gradients re-enter PyTorch's autograd graph zero-copy through DLPack.}
  \label{fig:training}
\end{figure*}

The decisive performance property (\Cref{fig:training}) is that compilation and verification happen once, on the first forward pass. Every later iteration reuses the cached circuit, executing only the weight update and the level-parallel forward/backward kernels. This is the mechanism behind the $2.74\times$ training speedup; the loss reduction across seeds confirms that gradients flow from circuit evaluation through the logic program back to the CNN, and the full timing breakdown is reported in \Cref{sec:eval}.

\subsection{Zero-copy interoperability}\label{subsec:interop}

A GPU-native engine is only useful if results reach the frameworks researchers already use, without copies that would reintroduce the transfer wall. xlog exposes its device-resident state through four standard interfaces.

\textbf{DLPack.} Each relation column or gradient buffer becomes a managed DLPack~\cite{dlpack} tensor whose GPU memory is reference-counted and freed only when every consumer releases it; on import, xlog validates device, dtype, contiguity, and alignment, optionally against an expected schema. In Python the protocol surfaces as a capsule, so \texttt{torch.from\_dlpack} yields a live CUDA tensor backed by the very memory xlog wrote.

\textbf{Arrow.} For columnar tools (Polars, DuckDB, cuDF), relations export through the Arrow~\cite{arrow} C Device Data Interface; symbol columns export as Arrow dictionary arrays, so a consumer reads symbolic columns as native string dictionaries while xlog keeps the compact integer representation.

\textbf{Python bindings.} The \texttt{pyxlog} extension (PyO3~\cite{pyo3}) exposes compilation, evaluation, training, and result extraction; tensor-valued results are returned as DLPack capsules, and long-running GPU operations release the GIL so data-loader threads proceed concurrently.

\textbf{PyTorch autograd.} The circuit behaves as a differentiable function inside PyTorch: the network's grad-enabled forward feeds the circuit, the scalar loss is exported via DLPack and re-imported, and a batched path stacks inputs into one forward pass and runs a single backward, so gradients flow from the logic layer back to \texttt{nn.Module} parameters with no custom autograd \texttt{Function}.

\subsection{Trainable rule bodies}\label{subsec:trainable-bodies}

Placing a neural predicate in a rule body lifts xlog from learning a single perceptual predicate to learning whole rules. The tensor and circuit data planes remain device-resident; route-specific orchestration and control observations are documented separately.

\textbf{Mixed bodies.} A trainable body may interleave neural-evaluated atoms with ordinary symbolic literals. The symbolic literals act as hard join conditions: they gate which groundings can fire but carry no probability mass or gradient. The head probability therefore factors as a $0/1$ indicator of the symbolic condition times the neural probability scaled by a learned per-rule weight~$\sigma(w)$, and gradients flow only through the network outputs and the rule weight, never through the ground facts.

\textbf{Existential joins.} When an existential (non-head) body variable is a neural predicate's input, xlog grounds that predicate over the \emph{real} join domain \emph{inside} the circuit instead of stripping the join as a pre-filter: the neural occurrence expands into one circuit leaf per domain event, and provenance OR-aggregates $\exists\,e.\ \mathrm{neural}(e)\wedge\mathrm{join}(e,\mathit{head})$ at each head binding. This makes ``does supporting evidence exist?'' rules learnable, reasoning over a whole domain of events instead of only per-instance classification.

\textbf{Joint multi-rule mixtures.} Several same-head trainable clauses combine as a noisy-OR mixture, $P(\mathit{head}_i)=1-\prod_k\bigl(1-r_k[i]\,m_k[i]\,\sigma(w_k)\bigr)$. Here $r_k[i]$ is a candidate's relational eligibility, $\sigma(w_k)$ its learned rule weight, and $m_k[i]$ an optional graded confidence in $[0,1]$ carried per binding, for example a fact's confidence trajectory over time. Rule learning thus moves from binary rule eligibility to a continuous, differentiable weighting of competing and cooperating rules.

\textbf{Learned body gates.} Inside that mixture a candidate may also carry a neural conjunct over a per-binding entity feature $\phi(x)$: its eligibility becomes its relational grounding \emph{and} a straight-through-thresholded gate $g_\theta(\phi(x))\ge\tau$. A gated clause then competes against ungated ones on the same noisy-OR terms, and $g_\theta$ trains jointly with the rule weights under the same held-out selector. By default $\phi$ is detached where it enters the engine: gradient reaches $\theta$ and the rule weight but stops at the neural head, leaving the feature producer frozen. Training the producer too is an explicit opt-in per candidate: one flag leaves $\phi$ attached, so gradient flows on into the backbone that computed it. Only the autograd linkage changes; the values uploaded are identical either way.

These trainable-rule paths share the verified circuit and the zero-copy substrate of the simpler predicate case, and extend to recursive programs now that provenance converges on two-sided recursive strongly-connected components. \Cref{sec:ecinduction} carries the surface onto a public benchmark: the perceptual predicate inside an induced Event-Calculus rule body is trained through symbolic credit alone, cross-validated under pre-registered gates. \Cref{sec:limits} states what that evidence does and does not support.

\subsection{Engine-mode credit and candidate selection}\label{subsec:engine-mode}

Which of the candidate bodies the engine enumerates should the learner keep, and on what evidence? Learning a rule body is not only a gradient problem: it is also a decision about which candidate to keep, and the two need different evidence. Engine-mode training separates them. The engine already enumerates the candidate space (the well-formed pairs of body relations over the program's own binary relations) and already holds the join extensions those candidates read, so training scores that space directly instead of a reconstruction of it.

A candidate's score on a fact is a noisy-OR over the witnesses the engine reports for that fact: a neural body relation contributes a probability per witness, read at the fact's own label, while a purely relational one contributes a fixed $\{0,1\}$ cover. Candidates combine as a softmax over their logits, and the resulting negative log-likelihood is a single autograd graph, so one backward pass updates the candidate logits and the network behind the probabilities together.

\textbf{A worked instance.} Take one head fact and two candidates for its body, one relational and one neural. The relational candidate covers the fact or does not, so its score there is $1$ or $0$. The neural candidate's score is the noisy-OR over the witnesses the engine reports for that fact, each read at the fact's own label: two witnesses at probability $0.5$ give $1-0.5\cdot0.5=0.75$, above a relational candidate that does not cover the fact and below one that does. Those are training scores, and the arbiter never reads them. It rescores both on the held-out facts by the same witness semantics, drops either whose held-out score falls below the fit gate, and keeps the relational one when the two land within the score quantum; if that narrowing leaves two relational candidates the data cannot separate, it abstains rather than pick one.

The candidate pool is validated against the typed registry of \Cref{subsec:neural-pred} instead of being trusted by name: a neural relation whose declared arity the credit has no witness semantics for is refused at registration, and a pool that filtering empties reports its per-filter counts instead of silently shrinking. And because raw engine constants index the caller's feature rows directly, that identity is accepted only when the witness domain is exactly the dense row range; a misalignment that would stay in bounds while gathering another event's probability draws a refusal instead of a guess.

\textbf{Selection on held-out generalization.} Training credit cannot distinguish a crisp but coincidental rule from a soft but correct one, even in principle; generalization can. Selection therefore never reads the trained weight. The facts are split into folds; each fold trains on the rest and rescores every candidate on the held-out facts by its own witness semantics; the arbiter ranks the fold-averaged scores. A \emph{fit gate} then drops any candidate whose held-out score falls below a threshold, since a rule that cannot fit held-out data is not a rule.

Ties within the score quantum break toward the relational candidate on Occam grounds, but only when that narrowing leaves exactly one: preferring a relational explanation over a neural one is licensed, choosing among relational duplicates the data cannot separate is not. When neither step is decisive the arbiter returns no rule and says why; abstention is one of the selector's outcomes. The held-out score may be accuracy or F1, and the choice matters: at a rare positive class accuracy sits on the all-negative base-rate plateau, where a candidate that predicts nothing outranks the best real detector, which is why the benchmark protocol of \Cref{sec:ecinduction} selects on held-out F1 instead.

\textbf{Masked witnesses and withheld evidence.} A per-witness mask lets evidence be withheld rather than denied: a masked witness leaves the candidate's index entirely, contributing zero credit and zero gradient instead of a coerced false, and the facts it thereby leaves uncertain are excluded from the held-out score and counted against a coverage gate that runs before the fit gate, so no candidate is killed by facts it was never allowed to see. The channel rides the accuracy axis only and is refused up front beside the F1 holdout, whose measurability derives from raw labels a mask can hide.

\textbf{Acceptance without retraining.} The same machinery also runs as an acceptance instrument: given an externally trained detector, a frozen entry point scores every enumerated candidate against it with no optimizer, no gradient and no folds and applies exactly the arbiter's gates, so where the cross-validated path asks whether a detector can be trained for a rule, this one asks whether a rule is right given the detector one already has. A training-mode module is refused rather than quietly switched, since batch-norm statistics and dropout would de-determinize the bit-identical scores the entry point exists to provide.

\subsection{Term embeddings and differentiable ILP}

\textbf{Term embeddings} attach dense, optionally trainable vectors to logical symbols. Where \texttt{nn/4} maps perception to discrete facts, embeddings give logical entities a continuous representation that participates in neural computation, with gradients flowing through the embedding lookup: knowledge-graph entity representations, for instance, trained jointly with neural perception.

\textbf{Differentiable ILP} learns first-order rules instead of network weights. Candidate rules are represented as sparse bitmasks over the ground-atom space, credit assignment runs entirely on-device via scatter--gather, and a promotion pipeline decides which rules enter the program.

Four kinds of judgement sit on these two paths, and \Cref{tab:gates} separates them. Convergence and the supply of explicit held-out examples are preconditions checked before any gate runs: with neither held-out positives nor held-out negatives supplied, the pipeline returns for manual review before a gate is evaluated, so the always-run gates cannot promote on their own.

The benchmark runs of \Cref{sec:ecinduction} exercise the last row only. Its permutation-null threshold is not an extra gate but the holdout arbiter's own fit threshold, derived per fold from label permutations instead of fixed at a constant; the promotion gates above it belong to the differentiable-ILP path and are not on that route.

\begin{table}[htbp]
  \centering
  \caption{What judges what, on which path: the promotion pipeline of differentiable ILP or the holdout arbiter of \Cref{subsec:engine-mode}. Convergence and supplied held-out examples are preconditions, not gates: without them the pipeline returns for manual review before any row below runs.}
  \label{tab:gates}
  \footnotesize
  \setlength{\tabcolsep}{3pt}
  \begin{tabularx}{\columnwidth}{@{}>{\raggedright\arraybackslash}p{0.25\columnwidth}Ll@{}}
    \toprule
    \textbf{Judgement} & \textbf{What it judges} & \textbf{Path} \\
    \midrule
    Six always-run gates & the rule derives every training positive and no training negative, passes the novel-fact audit, loses no protected fact, meets a held-out F1 threshold, and names only relations with typed schemas & dILP \\
    \addlinespace
    Two held-out gates & the same rule on the supplied held-out positives and negatives & dILP \\
    \addlinespace
    Ambiguity scan & which other candidates also explain the data; reported, never decisive & dILP \\
    \addlinespace
    Permutation-null fit threshold & whether a candidate generalizes better than shuffled labels do on its own fold & arbiter \\
    \bottomrule
  \end{tabularx}
\end{table}

Both paths ride the device-resident bridge of \Cref{subsec:interop}. Where the trainable rule bodies of \Cref{subsec:trainable-bodies} learn rule \emph{weights} and per-binding confidences, differentiable ILP learns rule \emph{structure}; the two are complementary. This sparse, GPU-resident formulation avoids the cubic blowup of dense rule materialization. Unlike the trainable-body surface, this path is not exercised on the benchmark of \Cref{sec:ecinduction}, and it is a beta subsystem (\Cref{sec:limits}).

\section{Learning Event Rules from Perception}\label{sec:ecinduction}

The preceding sections build the machinery; this one puts it on an external benchmark. The task is symbolic event-rule induction over video surveillance data, and the perceptual predicate the induced rules depend on is itself learned through the logic credit alone, with no perceptual supervision anywhere in the loop.

The Event Calculus is the standard logical formalism for narrative reasoning over time: a \emph{fluent} is a time-varying property of the world, for instance that two tracked people are \texttt{meeting}, and the two event predicates \texttt{initiatedAt(F,T)} and \texttt{terminatedAt(F,T)} say when something at time \texttt{T} makes fluent \texttt{F} begin or makes it stop. A domain-independent axiom of inertia closes the loop, so that a fluent \emph{holds at} a time if it was initiated earlier and not terminated since. Reasoning is therefore cheap and generic once the initiation and termination rules are known; the whole difficulty is obtaining them.

Those rules are classically hand-written by a domain expert. The line of systems that learns them instead, inductive logic programming over Event-Calculus theories, learns them from a symbolic input vocabulary that somebody else has already computed. On the CAVIAR video benchmark the learner is handed a relation such as \texttt{close(P1,P2,T)}, already reduced from raw tracker coordinates by a hand-chosen distance threshold, alongside per-person activity states. Rule induction thus starts one layer above perception, and the perceptual layer's quality enters as a fixed input that the learning does not explain.

In the runs below the proximity predicate is not given. It is \texttt{close\_nn}, a small network over the raw pair coordinates, and it is never shown a distance label, a threshold, or any target of its own. Its only gradient arrives through the induced rules: a candidate clause containing \texttt{close\_nn} in its body is scored by how well the resulting Event-Calculus theory explains the frame-level annotations, and that score is what propagates back to the network's weights.

This is the trainable rule bodies of \Cref{subsec:trainable-bodies} on a real benchmark instead of a demonstration: the neural atom sits inside a rule body, and the symbolic literals beside it act as hard join conditions that gate which groundings fire without carrying gradient. These runs take the engine-mode credit path of \Cref{subsec:engine-mode}: the rule learner's own negative log-likelihood is the network's loss, so structure search and weight training are one differentiable object rather than two pipelines in sequence. The compiled route of \Cref{subsec:trainable-bodies,subsec:interop}, with its machine-checked equivalence certificate and its zero-copy DLPack export, is the one the MNIST-addition measurements of \Cref{sec:eval} exercise; \texttt{close\_nn} here is trained by a different route.

\subsection{Protocol}

Two targets are searched throughout this section. The \emph{direct} target asks, per timestep, whether the fluent holds; the \emph{Event-Calculus} target asks when it is initiated and terminated, and infers holding through inertia. They carry different class balances, and the guarantees below are stated per target. Except where a table says otherwise, F1 is micro-averaged over pooled per-fold counts, as the published evaluations compute it.

Each guarantee below was fixed before the run it governs and covers that run alone: the corpus was already in hand while the gates were being designed.

\textbf{Leak-free scene-family splitting.} The distributed CAVIAR corpus used by the published Event-Calculus learners is a dump of 26 video segments, and it is not clean. A frame-level audit of every \texttt{meeting} transition event in the dump against the original ground-truth XML classifies 21 of 25 events as real, 3 as duplicates and 1 as a splice artifact. Two source videos appear in both the dump's train and its test file, and one video alone contributes 855 duplicated gold pair-frames. Cross-validating over segments therefore permits same-scene transfer. The clean protocol folds over \emph{scene families} recovered from the source XML instead, so no scene appears on both sides of a split.

\textbf{Recall-aware holdout.} Scoring candidates by held-out accuracy is useless at this class balance (roughly ten positive transitions against tens of thousands of rows), because every candidate sits on the all-false base-rate plateau, and the ranking provably inverts: the best real detector scores below the empty theory. Candidate selection in the Event-Calculus searches uses per-fold held-out F1 instead. The direct per-timestep target does not have this problem, since it carries thousands of positives per fold and not a handful of transition events, so its searches are left at their defaults, held-out accuracy with a fixed $0.75$ fit gate. This guard and the next therefore govern the Event-Calculus route only.

\textbf{Permutation-null fit gate.} A candidate must beat what label shuffling alone achieves on its own fold. Each fold derives its own threshold from 1000 label permutations (seed 7) as the 95th percentile of the pool-maximum mean per-fold F1, computed per target and on the training side only. A theory whose clauses do not clear that bar is not returned: the search abstains and reports an empty theory. Abstention is a first-class outcome of this protocol.

\textbf{Two fold structures.} The direct per-timestep runs reuse the three folds the distributed dataset file ships, whose datapoints are fixed-length windows of consecutive frames for one person pair (the \emph{windowed folds}); the Event-Calculus runs fold over whole video segments, or over scene families under the clean protocol.

\subsection{Results}

\textbf{Perception learned through logic credit.} On the windowed-fold protocol with a direct per-timestep target, the search finds the same two-clause theory on every fold in both modes (\Cref{lst:ec-theory}):

\begin{lstlisting}[style=xlogstyle, basicstyle=\ttfamily\scriptsize,
                   caption={The two-clause theory the search returns on every
                            fold, in both the relational and the neural mode.},
                   label={lst:ec-theory}]
holdsAt_meeting(PP,T) :- both_inactive(PP,T), close*(PP,T).
holdsAt_meeting(PP,T) :- both_active(PP,T),   close*(PP,T).
\end{lstlisting}

\noindent where \texttt{close*} is the precomputed geometric predicate in the relational mode and the learned \texttt{close\_nn} in the neural mode. Held-out F1 per fold is $0.9215 / 0.6804 / 0.7695$ with the precomputed predicate (mean $0.7905$) and $0.9215 / 0.7918 / 0.7695$ with the learned one (mean $0.8276$). The learned detector matches the precomputed predicate's result exactly on two folds and outperforms it on the third, where a soft learned boundary survives the train/test geometry shift better than a hard threshold. That is this section's main positive result: a perceptual predicate trained through nothing but symbolic credit stands in for hand-set geometry without loss, and rests on no published comparison.

\textbf{Protocol-matched cross-validation.} Under the full Event-Calculus target, the combined 26-segment corpus (32{,}360 co-visible pair-frames, 1{,}833 gold \texttt{meeting} frames) was cross-validated ten-fold by video segment with every gate re-derived inside each fold. The result is precision $0.6580$, recall $0.8271$, F1 $0.7329$. The same initiation clause, \texttt{both\_active \& close}, is selected on all ten folds, each time over that fold's own permutation null; the termination theory is empty on all ten, so the fluent persists by inertia to the end of each co-visible pair-run.

\textbf{Published figures on the same corpus.} This protocol matches the published Event-Calculus learners on the axis that matters most for the metric: F1 is computed on \texttt{holdsAt} inferred through inertia, on the same distributed corpus, with the same micro-aggregation. The published figures for \texttt{meeting} on that corpus are $0.735$ for hand-crafted rules, $0.782$ and $0.792$ for OLED~\cite{oled} (the latter as reported by its authors, the former as re-run in the WOLED paper~\cite{woled}) and $0.887$ for WOLED-ASP; they are tabulated beside ours in \Cref{sec:eval}. Ours sits just under that group's lowest figure. We draw no ordering from any of these differences, in either direction. Five protocol deltas separate the rows and they do not all point the same way: the fold structure, the scope of the learned theory, the body-length limit and the way the reported settings were chosen all differ, and the shared metric is the only reason the comparison is drawn at all. \Cref{tab:ec-published} prints the five deltas in full beside the figures.

\textbf{A second vocabulary iteration.} Two pair-level transition relations were added to the Event-Calculus vocabulary, one for a pair crossing the proximity threshold outward and one for strictly growing distance, and the identical folds, seeds and gates were re-run. The result rises to precision $0.7357$, recall $0.8260$, F1 $0.7782$ (1514 true positives, 544 false positives, 319 false negatives), with a termination theory now learned on seven of the ten folds. That figure falls inside the $0.735$--$0.887$ span the published systems occupy, but it must never be read as the headline: its vocabulary was chosen \emph{after} the first run's test-side failure mode was known, on the same folds. Everything else about it is pre-registered; the vocabulary is not. Against published estimates selected as the best of several tried, the smaller number declared in advance is the stronger scientific claim, which is why it leads here and $0.7782$ appears only with this label attached.

\textbf{The learned detector under the Event-Calculus target.} Substituting the learned \texttt{close\_nn} detector for the precomputed proximity predicate inside the Event-Calculus initiation search does not survive cross-validation: precision $0.1250$, recall $0.0005$, F1 $0.0011$ over the same ten folds (\Cref{tab:ec-runs} gives the counts). Only one fold commits a clause, the same \texttt{both\_active \& close\_nn} shape the geometric predicate selects there, and probed directly against held-out ground-truth proximity that clause scores precision $1.0$ at recall $0.27$, so the network has learned the geometric relation. The failure is one of scale in the selection rule, not of perception: the initiation search scores coverage against transition \emph{events}, a handful per fold, and a probability-thresholded gate newly covers fewer of them than a deterministic predicate does, so nine of ten folds reject the candidate for insufficient new coverage.

\textbf{The leak-free protocol.} Re-run on the deduplicated, scene-family-grouped corpus, the \texttt{meeting} fluent yields F1 $0.0$ on both routes, and mostly by committing a rule that earns nothing rather than by declining to commit one. The direct route selects \texttt{both\_inactive \& close} on nine of the ten folds, and that clause returns no true positive anywhere (0 tp, 106 fp, 1{,}812 fn). The Event-Calculus route commits \texttt{both\_active \& close} on one fold, whose 120 false positives are its entire error mass (0/120/1{,}812), and abstains on the other nine, recording insufficient new coverage or an abstaining selector; its termination search returns nothing on any fold, most often because no body passed that fold's fit gate. Both routes produce false positives, which an empty theory cannot: the search ran, and what failed is transfer.

The reason is visible in the data. One scene family carries 1{,}323 of the 1{,}812 \texttt{meeting}-positive frames and, under honest grouping, sits in exactly one fold; the clause that explains those frames holds \emph{only} there, covering zero of the 489 positives on all eight remaining \texttt{meeting}-bearing segments. Both outcomes follow from that. Where the clause is committed, it is committed on a training side that contains the concentrated family, and it covers nothing outside it. The one fold on which the direct route abstains is precisely the fold that holds that family out, leaving a training side on which the clause covers 0 of 489 positives and 106 negatives.

Where the gate abstains, it is because the candidate's held-out score has collapsed for the same reason. The statistical bar itself did not rise: the per-fold null thresholds under this protocol are essentially those of the like-for-like distributed-corpus run. Training with the family learns a clause that transfers nowhere; training without it cannot learn the clause at all. With eleven observable \texttt{meeting} initiations and five \texttt{moving} ones, the clean corpus does not carry enough evidence for a cross-validated theory in either direction.

The machinery is not broken under this protocol. On \texttt{moving} the direct route still selects the canonical published rule shape on nine of ten folds (micro F1 $0.4450$), and the \texttt{meeting} clause applied to the concentrated family's own fold, an in-family fit and not a transfer test, scores frame F1 $0.890$ there. What it establishes is narrower and more specific than an abstention: a two-literal state vocabulary does not carry CAVIAR \texttt{meeting} across scene families at all.

\begin{table}[t]
  \centering
  \small
  \setlength{\tabcolsep}{4pt}
  \begin{tabularx}{\columnwidth}{Lccc}
    \toprule
    \textbf{Run} & \textbf{P} & \textbf{R} & \textbf{F1}\\
    \midrule
    \multicolumn{4}{l}{\emph{Windowed folds, direct target (mean of 3 folds)}}\\
    \quad precomputed \texttt{close} & --- & --- & 0.7905\\
    \quad learned \texttt{close\_nn} & --- & --- & \textbf{0.8276}\\
    \addlinespace
    \multicolumn{4}{l}{\emph{Distributed corpus, 10-fold CV, EC + inertia}}\\
    \quad pre-registered & 0.6580 & 0.8271 & \textbf{0.7329}\\
    \quad + termination relations & 0.7357 & 0.8260 & 0.7782\\
    \quad learned \texttt{close\_nn} & 0.1250 & 0.0005 & 0.0011\\
    \addlinespace
    \multicolumn{4}{l}{\emph{Clean protocol (scene-family folds)}}\\
    \quad \texttt{meeting}, EC route & 0.0 & 0.0 & 0.0\\
    \quad \texttt{meeting}, direct route & 0.0 & 0.0 & 0.0\\
    \quad \texttt{moving}, EC route & 0.0 & 0.0 & 0.0\\
    \quad \texttt{moving}, direct route & 0.5334 & 0.3817 & 0.4450\\
    \bottomrule
  \end{tabularx}
  \caption{Runs reported in this section. Micro F1 over pooled per-fold counts, except the windowed rows, which are fold means. The \emph{+ termination relations} row is the second, adaptive vocabulary iteration. The clean-protocol zeros are scored predictions, not empty theories, except the \texttt{moving} Event-Calculus row, whose initiation theory is empty on every fold; the census behind them is in the text. Published figures for the same benchmark are tabulated separately in \Cref{sec:eval}.}
  \label{tab:ecinduction}
\end{table}

\subsection{Discussion}

Two of the findings above concern the method. A perceptual predicate can be learned end-to-end through symbolic credit alone and stand in for hand-set geometry without loss on a real benchmark. And Event-Calculus rules can be induced under fully pre-registered gates and land within a couple of points of the range published Event-Calculus learners report on the corpus they share.

The other two concern that corpus. It carries a measurable duplication defect, and under a leak-free split of it the induced \texttt{meeting} rule does not transfer: the search still commits the canonical clause on most folds, and that clause scores zero out of family.

None of this ranks the system against the published learners (\Cref{sec:limits}). Termination programs are not learned in the pre-registered run either, and the empty termination theory is the dominant source of false positives there. Nor is the learned detector a drop-in replacement for the geometric predicate everywhere: under Event-Calculus cross-validation it is not. And it is one benchmark in one domain. What the section does show is that the structure search and the perception underneath it can be one differentiable object instead of two systems in sequence.

\section{Rule Induction at Scale:\texorpdfstring{\\}{ }Maritime Rendezvous}\label{sec:maritime}

The previous section ends on an evidence problem: under a leak-free split CAVIAR carries eleven observable \texttt{meeting} initiations, too few for a cross-validated claim about rule learning. This section moves the same machinery to the other public corpus the Event-Calculus learners report on, the Brest AIS maritime stream~\cite{brestais,maritimecer}, whose target fluent \texttt{rendezVous} (two vessels close together, stopped or moving slowly, in open water) carries 3{,}548 gold intervals.

It is large enough to separate three questions CAVIAR could not. Does a \emph{crisp} rule search, selecting clauses that each fire with weight one as in \Cref{sec:ecinduction}, land where the declared ceiling says it will? How much is gained by \emph{weighting} the clauses of the same gated pool instead of selecting them? And what does it cost to learn those weights in one chronological pass over the stream? The six runs of \Cref{tab:maritime} answer them, covering the published systems' three training settings: enumerated rule search, weighted clauses, single-pass weight training.

Two of the three answers came out better than we expected.

\subsection{Corpus and labels}

The inputs are two public exports of RTEC~\cite{rtec}, a hand-written Event-Calculus rule engine, over the Brest AIS stream: the composite events it recognised (Zenodo record 2557290) and its own input, the \emph{critical points} of the compressed trajectories~\cite{maritimecer}, the timestamps at which a vessel stops, turns, changes speed or loses signal. Those points are the only temporal grid the corpus carries. A deterministic converter reduces the two into a \emph{pair-time corpus}: one row per (vessel pair, timestamp), eleven relations that hold or do not for that pair then, and the gold label. \Cref{tab:maritime-corpus} sizes it. The conversion uses no randomness: the corpus is a pure function of the archives.

\begin{table}[htbp]
  \centering
  \caption{The pair-time corpus, as the converter leaves it. The negative pairs are drawn by fixed stride from the $2{,}014$ pairs that come within proximity but carry no interval.}
  \label{tab:maritime-corpus}
  \small
  \begin{tabularx}{\columnwidth}{@{}Lr@{}}
    \toprule
    \textbf{Quantity} & \textbf{Value} \\
    \midrule
    Rows (vessel pair $\times$ timestamp)          & $454{,}858$ \\
    Vessel pairs                                   & $806$ \\
    \quad carrying a \texttt{rendezVous} interval  & $302$ \\
    \quad negatives, by fixed stride               & $504$ \\
    Gold intervals                                 & $3{,}548$ \\
    Positive rows                                  & $3{,}579$ \\
    \quad as a share of all rows                   & $0.79\%$ \\
    \bottomrule
  \end{tabularx}
\end{table}

Three properties bound what a number measured on it can mean. The supervision is synthetic: the gold labels are a hand-crafted RTEC rule's output over the same interval streams the vocabulary comes from, so their perfect alignment with those streams says the conversion is faithful, not that the labels are right. The published learners share this regime; the WOLED paper notes that the maritime ground truth is synthetic and its learning curves consequently alike~\cite{woled}. What every system here measures is the \emph{reconstruction of a known rule} under a vocabulary that does not contain all of it.

The vocabulary is incomplete by construction. Its eleven pair-time relations (\texttt{proximity}, \texttt{both\_lowspeed}, \texttt{both\_stopped\_far}, \texttt{both\_open\_sea}, \texttt{became\_proximate}, and so on) omit the gold rule's two remaining discriminators, its $240$\,s minimum-duration threshold and its exclusion of tug and pilot pairs, so perfect reconstruction in the base vocabulary is impossible.

And the grid is sparse: ${\approx}1.009$ positive rows per gold interval, the covering row almost always the interval's start boundary, so pointwise and interval F1 coincide here (\Cref{subsec:maritime-comparison}).

\subsection{Protocol}

\textbf{Metric.} The primary metric is pointwise F1 on the positive class, F1 over individual rows, never accuracy, which the empty predictor saturates here. Interval figures accompany it, a gold interval counting as matched when a predicted row falls inside it in the same episode of the pair's timeline, and symmetrically.

\textbf{The declared ceiling.} Each run's hypotheses and parameters were fixed before it was executed, beginning with the ceiling the base vocabulary imposes. The gold rule's definitional body is \texttt{proximity \& both\_low\_or\_stopped \& both\_open\_sea}, the middle literal being the disjunction of \texttt{both\_lowspeed} and \texttt{both\_stopped\_far}; all three are in the vocabulary, so the body is expressible either way. It reaches recall $1.0$ at pointwise precision ${\approx}0.49$, hence F1 ${\approx}0.66$: the two inexpressible discriminators are exactly what separate its false positives from its true ones.

A direct evaluation of that body on the archives, the \emph{ceiling probe}, hits the operating point exactly: F1 $0.6599$, no false negative (\Cref{tab:maritime-folds}). It also decomposes them: $733$ in negative pairs, the pair-exclusion class, $2{,}956$ in predicted runs shorter than $240$\,s, and \emph{none} outside the two pre-identified discriminators.

\textbf{Folds are vessel pairs.} A pair's rows are never split across folds, because the positive mass is concentrated: the top pair carries $33.4\%$ of all positive rows and the top four $60.2\%$, so a finer split would leak the dominant encounter across the boundary. The protocol uses five folds where CAVIAR uses ten, assigned by a deterministic greedy longest-processing-time rule with no seed, stratified by positive mass, none differing from another by more than the largest pair's count. The consequence is stated openly: held-out fold 0 is the top pair plus one negative pair, so per-fold figures are necessarily unequal and the micro number never stands alone.

\textbf{Search and gate as on CAVIAR.} The crisp run uses the relational sequential-covering search of \Cref{sec:ecinduction} under the CAVIAR limits: bodies of at most three literals, at most four clauses, per-fold held-out F1 for candidate selection, at least two newly covered positives per clause. Its permutation-null fit gate is the CAVIAR one: $1{,}000$ label permutations per fold, threshold at the $95$th percentile of the pool-maximum per-fold F1, derived on the training side. One seed, $7$, drives the permutations and the candidate-selection holdout. The target is the direct per-row positive label; induction through inertia is out of scope here, with almost no interior rows to fill.

\textbf{Weights.} The weighted runs keep vocabulary, folds and gate, replacing crisp selection with weights. The body pool is every conjunction of one to three literals passing the same per-fold permutation-null procedure, thresholds re-derived per run, each body carrying one weight. A row's score is the noisy-OR $1-\prod_c\bigl(1-\sigma(w_c)\,\mathrm{cover}_c\bigr)$ of \Cref{subsec:trainable-bodies}, every clause purely relational, trained by binary cross-entropy with Adam and thresholded at $0.5$: $300$ steps at learning rate $0.05$ from every weight at $-2.0$, so every clause starts switched off ($\sigma(-2)\approx0.12$). The single-pass variants replace that optimisation with \emph{one} pass over the fold in ascending global time, its pairs interleaved as in a stream, in mini-batches of $1{,}000$ rows, one Adam step per batch, at the same rate. Adam is kept instead of WOLED's AdaGrad, so exactly one variable separates the regimes. Pool and gate are still computed on the training side, which makes this single-pass \emph{weight} learning, semi-online in that only the weights see the stream; the structure is never learned online.

\subsection{Results}

\textbf{Crisp search on the base vocabulary.} The first run is the CAVIAR search unchanged. It lands where the ceiling mechanism predicted, precision near one half at recall near one, for micro pointwise F1 $\mathbf{0.6746}$ (\Cref{tab:maritime}).

On every fold it commits both disjunctive branches of the gold body: \texttt{both\_open\_sea \&\allowbreak\ both\_stopped\_far \&\allowbreak\ proximity} and \texttt{both\_lowspeed \&\allowbreak\ both\_open\_sea \&\allowbreak\ proximity}. Three folds add a further clause, on two of them the subsuming disjunction relation itself. No clause is ever accepted below its fold's null gate.

The learned operator is not literally the definitional body, its precision being $0.5109$ against the body's $0.4924$, so the evidence is about the mechanism, not operator identity. The per-fold spread of \Cref{tab:maritime-folds} is the pair-concentration effect: the fold with the top pair's long clean encounters scores highest, while folds of many small pairs concentrate the duration-threshold false positives. Improvement past this run has to come from expressiveness rather than tuning.

\textbf{Weighted clauses on the same pool.} The second run changes one thing: the pool's clauses are weighted rather than selected. We expected it to stay within $[0.66, 0.70]$, the ceiling being a property of the vocabulary.

The hypothesis was rejected in the favourable direction. Micro F1 is $\mathbf{0.7398}$, $+0.0652$ over the crisp run on the same vocabulary, folds and gate, at markedly higher precision and lower recall; the abstract and \Cref{sec:intro} round that gain to $0.065$.

The expectation was wrong for an instructive reason: the $0.66$ ceiling is the F1 of the definitional body's operating point, crisp reconstruction at recall near one, and weights are not confined to it. Training moved the operating point to high precision at lower recall, which carries more F1.

Fold by fold the picture is mixed. The weighted run wins the micro figure and the median but only two of the five folds, and its weakest fold is worse than the crisp run's. That weights over the same clauses beat crisp selection of them is the gain the published line attributes to weighting~\cite{woled}.

\textbf{Both runs again with the duration relation.} The third and fourth runs add one relation, \texttt{sustained\_240}: a row carries it when $\texttt{proximity}\cap\texttt{both\_low\_or\_stopped}\cap\texttt{both\_open\_sea}$ holds continuously through a stretch of at least $240$\,s containing it. The constant is the gold generator's own, so the addition is language parity with RTEC, not a look at the data.

The ceiling probe was re-run on the widened vocabulary before any cross-validated run, against an expectation of ${\approx}0.907$ on the assumption that duration removes only the short-run false positives. The ceiling came out higher, at F1 $0.9969$, again with no false negative.

Duration also eliminates at least $711$ of the $733$ negative-pair false positives, the excluded pairs' body-condition episodes being almost always shorter than $240$\,s. The bound is a lower one because the $22$ false positives the widened probe still makes are not decomposed by class. Two consequences were predicted from that: the pair exclusion is largely redundant with duration here, and the duration vocabulary saturates, so the informative weighted-against-crisp comparison is the base-vocabulary one.

Both held. Crisp search on the widened vocabulary lands exactly on the derived ceiling and the weighted run equals it on every fold, so the directional hypothesis, weighted no worse than crisp on the median, holds trivially, with no headroom for weights.

\textbf{Weighted clauses learned in a single pass.} The last two runs change only how the second run's weights are trained, replacing batch optimisation with a single chronological pass. They use the base vocabulary only: \texttt{sustained\_240} at time $t$ depends on how long the current interval will last, a leak in a streaming setting, and was excluded before the run.

Three hypotheses were fixed beforehand: the single pass should lose at most $0.04$ micro F1 against the batch run, with anything below $0.70$ a negative result; a \emph{reverse} chronological pass should land within $0.02$ of the forward one; and the prequential error, scored on each batch before that batch is trained on, should fall along the stream.

The measured degradation is zero. The single pass reproduces the batch run's thresholded predictions exactly: the same counts on every fold and, as re-scoring with the stored weights confirms, the same decision on every row, at the same micro F1. The reverse pass is identical as well, and the prequential error falls from the first half of the stream to the second on all five folds in both orders (forward, fold 0: $0.0078\to0.0036$). Halves are the one criterion of these runs settled as the results were written up instead of before them.

Identical predictions are a statement about the decision structure, not evidence that the single-pass path silently fell through to the batch one: the weights themselves differ substantially between the three regimes (\Cref{tab:maritime}). The mechanism is simpler than robustness to the training regime. In all fifteen fold-by-regime cases exactly one body has $\sigma(w)>0.5$, always \texttt{became\_proximate \&\allowbreak\ both\_open\_sea \&\allowbreak\ both\_stopped\_far}, and the second-ranked body stays between $0.15$ and $0.46$, so the noisy-OR of the others never reaches the threshold on any held-out row. All three trainings reduce to the same crisp rule, and the predictions agree because that rule does.

The agreement is fragile. The highest score of a row \emph{not} covered by the top body reaches $0.493$ in fold 1 under the batch weights, $0.007$ below the threshold.

That fold has $1{,}270$ rows between $0.3$ and $0.5$: a small change of learning rate, step count or initialisation could push the second body over and move the counts by hundreds of rows. The declared scope is narrow. Prediction identity holds for this pool, corpus and threshold, hangs on the discrete structure of a thresholded decision, and is not a theorem.

\begin{table}[t]
  \centering
  \small
  \setlength{\tabcolsep}{2pt}
  \begin{tabularx}{\columnwidth}{Lcccc}
    \toprule
    \textbf{Run} & \textbf{P} & \textbf{R} & \textbf{F1} & \multicolumn{1}{c}{\shortstack[c]{\textbf{F1, fold}\\\textbf{median}}}\\
    \midrule
    \multicolumn{5}{l}{\emph{Base vocabulary (11 relations; declared ceiling ${\approx}0.66$)}}\\
    \hspace{0.6em}crisp search & 0.5109 & 0.9925 & 0.6746 & 0.6596\\
    \hspace{0.6em}weighted, batch & 0.8715 & 0.6426 & \textbf{0.7398} & 0.6928\\
    \hspace{0.6em}weighted, forward pass & 0.8715 & 0.6426 & 0.7398 & 0.6928\\
    \hspace{0.6em}weighted, reverse pass & 0.8715 & 0.6426 & 0.7398 & 0.6928\\
    \addlinespace
    \multicolumn{5}{l}{\emph{Duration vocabulary ($+$\texttt{sustained\_240}; ceiling $0.9969$)}}\\
    \hspace{0.6em}crisp search & 0.9939 & 1.0 & 0.9969 & 0.9950\\
    \hspace{0.6em}weighted, batch & 0.9939 & 1.0 & 0.9969 & 0.9950\\
    \bottomrule
  \end{tabularx}
  \caption{Maritime \texttt{rendezVous}: five-fold cross-validation over whole vessel pairs, seed 7. P, R and F1 are pointwise micro-averages over pooled per-fold counts; the last column is the median of the five per-fold values. The three base-vocabulary weighted rows are identical row by row, though their weights are not: fold 0's top body carries $\sigma(w)=0.6476$ in batch, $0.5880$ in the forward pass and $0.8398$ in the reverse, and the $L_\infty$ distance between the logit vectors of any two regimes is $1.06$--$2.95$. Interval-level F1: 0.6772 (crisp, base; 3{,}545 of the 3{,}548 gold intervals matched), 0.7435 (weighted, base), 0.9970 (both duration rows). No published figure appears here (\Cref{subsec:maritime-comparison}).}
  \label{tab:maritime}
\end{table}

\begin{table}[t]
  \centering
  \small
  \setlength{\tabcolsep}{2pt}
  \begin{tabular}{@{}lccccc@{}}
    \toprule
    \textbf{Base vocabulary} & \textbf{f0} & \textbf{f1} & \textbf{f2} & \textbf{f3} & \textbf{f4}\\
    \midrule
    crisp search & 0.9933 & 0.4690 & 0.6685 & 0.5699 & 0.6596\\
    weighted, all three & 0.9346 & 0.4237 & 0.8000 & 0.6928 & 0.5342\\
    \midrule
    \textbf{Pooled counts} & \multicolumn{5}{l}{tp / fp / fn}\\
    \midrule
    ceiling probe, base & \multicolumn{5}{l}{3{,}579 / 3{,}689 / 0}\\
    crisp search & \multicolumn{5}{l}{3{,}552 / 3{,}400 / 27}\\
    weighted, all three & \multicolumn{5}{l}{2{,}300 / 339 / 1{,}279}\\
    ceiling probe, duration & \multicolumn{5}{l}{3{,}579 / 22 / 0}\\
    \bottomrule
  \end{tabular}
  \caption{Per-fold F1 on the base vocabulary and the pooled counts behind \Cref{tab:maritime}. The three weighted regimes share a row because their predictions are identical. The probe rows evaluate the gold rule's definitional body directly; the duration probe's $22$ residual false positives are what its $0.9969$ ceiling reflects.}
  \label{tab:maritime-folds}
\end{table}

\subsection{Discussion}\label{subsec:maritime-comparison}

The published figure for \texttt{rendezVous} on this corpus is $0.98$ for OLED and both variants of WOLED, MLN- and ASP-based alike, with theories of eighteen literals, measured on a train/test split of halves~\cite{woled}. A second published experiment, on a six-sequence fragment, compares WOLED-ASP with batch learners and reports figures from $0.91$ to $0.97$ in a bar chart. None of them stands beside the rows of \Cref{tab:maritime}, because the two sets of numbers are not on one axis.

Theirs live on a dense grid, the critical points of every vessel with inertia filling each interval's interior. Ours live on the sparse per-pair grid the converter produces, and we do not reproduce the published per-timepoint reading on it. A split of halves is not a pair-level five-fold cross-validation either, and for the same reason we do not compare pass times with published training times.

What is comparable is comparable in kind only. The published systems saturate on their preprocessing of the same stream, and ours too once the duration discriminator is expressible, at the declared ceiling. And the gain the published line reports between OLED and WOLED on CAVIAR, and attributes to weighting, is reproduced here as $+0.0652$ micro F1 with everything else held fixed, on a corpus with $3{,}548$ gold intervals instead of eleven events, and again without loss when the weights are learned in one pass.

Within those bounds the findings are concrete. The crisp search reconstructs the gold body on every fold of a corpus two orders of magnitude richer in positive events, and lands on the declared ceiling. Weighting the same pool adds $0.0652$ micro F1 and $0.0332$ to the median against the expectation declared for it, on two folds of five. Duration saturates both search modes and makes the pair exclusion largely redundant. And single-pass weights reproduce the batch predictions exactly, for the reason named above.

No ranking against OLED or WOLED on \texttt{rendezVous} follows from any of it (\Cref{sec:limits}). Nor does it say anything about detecting rendezvous in the wild: every number here is a rule-reconstruction figure. It does not demonstrate online structure learning, the pool and gate being computed on the training side, nor a general robustness of thresholded noisy-OR predictors to the training regime. And it exercises none of the GPU substrate: every run here is a deterministic CPU procedure.

What it does show is that the induction protocol of \Cref{sec:ecinduction}, its declared gates, pair-level folds and permutation nulls, carries intact to a corpus where its verdicts are measurements and not anecdotes. There, weighted clauses and single-pass training, the two mechanisms the published line rests on, are measured in isolation and behave as their authors report.

\section{Epistemic and Other Reasoning Modes}\label{sec:epistemic}

The substrate generalizes beyond what is \emph{true} or \emph{probable}. This section covers three further modes that reuse the same CUDA join, circuit, and CDCL machinery (epistemic reasoning over world views, probabilistic aggregates, and well-founded semantics) broadening what a neural network can be integrated with while keeping data-plane execution GPU-backed.

\subsection{Modal surface and the epistemic IR}

Epistemic logic programming asks what is \emph{known} or \emph{possible} across the multiple stable models a program admits---the natural setting for introspection and reasoning under incomplete knowledge. A program may use four modal literals over a predicate: \texttt{know}, \texttt{possible}, \texttt{not know}, \texttt{not possible}. Instead of rewriting them away at parse time, the compiler preserves them in an Epistemic IR (EIR) between the AST and execution planning. The high-level dispatcher classifies modal dependencies before selecting Generate--Propagate--Test, a mode-specific positive-cycle fixpoint, or GPU-backed WFS. The accepted surface spans finite nested modal chains, epistemic integrity constraints (which prune candidate world views on the GPU), rules that couple several epistemic predicates, same-name multi-arity modal predicates, and recursive epistemic execution.

\subsection{World-view semantics}

A world view is a non-empty set of stable models. \texttt{know p} holds when \texttt{p} appears in every model of the world view; \texttt{possible p} holds when it appears in at least one. Two semantics are selectable. Under the default FAEEL (Founded Autoepistemic Equilibrium Logic), a circular derivation contributes a tuple only when independent founded support exists. A program containing only \texttt{p :- possible p.} therefore executes to an empty founded extension instead of failing with an unsupported-construct diagnostic. A Gelfond-1991-style compatibility mode admits that self-supporting tuple. Non-epistemic programs evaluate identically under both.

\subsection{Epistemic execution routes}\label{subsec:epistemic-gpu}

After EIR construction, acyclic modal programs follow a Generate--Propagate--Test schedule whose phases run as CUDA kernels. \emph{Generate} enumerates bounded candidate assumptions as device bitsets; \emph{propagate} prunes candidates that contradict known or rejected facts; \emph{test} validates survivors against the program's world views, checking stable-model tuple membership on-device per arity. Validating a candidate on this route reuses the on-GPU CDCL solver already built for compilation verification (\Cref{sec:prob}), the same substrate that serves satisfiability, bounded MaxSAT and assumption-based queries, and accepted world-view evidence feeds the existing GPU-native exact path.

A positive FAEEL dependency cycle reduces to ordinary rules and reaches the founded least fixpoint in the existing recursive engine. A supported Gelfond-1991 cycle, whose selected positive \texttt{possible} gates must use their rule heads' own terms and stay within one recursive dependency component, instead descends to the greatest compatible set: the runtime relaxes those gates for an upper bound and then reevaluates until GPU set comparison over the intensional relations reports no further change (\Cref{sec:limits}). Supported cycles through negation take the GPU-backed WFS alternating fixpoint instead, and recursive negation or aggregation inside a compatibility component is rejected, because either would make the descending operator nonmonotone.

The reduced ordinary work reuses the optimizer, WCOJ planner and helper-splitting machinery of \Cref{sec:datalog}, and \emph{epistemic splitting} decomposes eligible acyclic programs into independent components solved separately and recombined. No CPU fallback is provided for the accepted data-plane hot paths: a construct these routes do not support is rejected with a typed error; no host implementation serves it. Transfer budgets track data movement; the host sequences Gelfond-1991 refinement and WFS iterations, while relation evaluation, snapshot copies and convergence comparisons remain GPU-backed.

\subsection{Probabilistic aggregates}

Finite aggregate queries (\texttt{count}, \texttt{sum}, \texttt{min}, \texttt{max}, \texttt{logsumexp}) are supported in both exact provenance/PIR inference and Monte Carlo execution, subject to a typed exact-domain cap that rejects domains too large for tractable enumeration. For small finite \texttt{count} domains, \emph{aggregate lifting} replaces naive world enumeration with exact cardinality dynamic programming over a compact state set (collapsing, for example, a $2^{17}$-outcome enumeration into a few hundred lifted states with identical results) and accepted \texttt{count} aggregates dispatch a GPU-native exact evaluator rather than a CPU finite-world shortcut, keeping the path device-resident.

\subsection{Well-founded semantics}

Programs that violate stratification, such as the classic \texttt{p :- not q.\ q :- not p.}, do not have a unique two-valued least-fixpoint interpretation. For supported shapes, xlog instead assigns three-valued truth (true, false, undefined) through well-founded semantics, computing an alternating fixed point: mark the greatest unfounded set false, derive consequences true, and repeat until stable. For probabilistic programs, true atoms receive normal probability and gradient while false and undefined atoms are assigned probability zero, matching the conservative treatment of non-stratified programs. Probabilistic provenance analysis and epistemic dependency dispatch select the GPU-backed WFS route for the nonmonotone components they support; other cyclic shapes return a typed diagnostic.

\section{Evaluation}\label{sec:eval}

This section reports measurements for xlog's deterministic, probabilistic, and neural-symbolic subsystems. The subsystem measurements in \Cref{sec:wcoj-eval,sec:cache-eval,sec:runtime-eval} are \emph{single-system ablations of xlog against its own baselines}: the first two on development hardware, the runtime-optimization pair on a cloud GPU that the artifact names. \Cref{sec:h2h} then adds controlled head-to-head comparisons against external engines (Scallop, ProbLog2, Soufflé), \Cref{sec:ec-eval} tabulates the Event-Calculus rule-induction runs of \Cref{sec:ecinduction} beside the figures published for the same benchmark and states why the maritime runs of \Cref{sec:maritime} are not tabulated beside theirs, and \Cref{sec:overhead} isolates xlog's two design costs, verification and residency, from ordinary GPU acceleration. \Cref{sec:noquant} describes three further runtime capabilities for which no timing is claimed.

\subsection{Methodology}

The subsystem ablations were collected on an NVIDIA RTX PRO 3000 (Blackwell, 12\,GB VRAM, compute capability 12.0). The head-to-head comparisons of \Cref{sec:h2h}, the overhead isolations of \Cref{sec:overhead} and the runtime-optimization ablations of \Cref{sec:runtime-eval} ran on cloud GPUs, as stated with each, and the neural Event-Calculus runs of \Cref{sec:ec-eval} on an NVIDIA A40. Each cloud artifact records its own device, driver and CPU quota. For the seven head-to-head and overhead artifacts, a number that differs from an earlier version of the same artifact differs because of the machine: those seven were all measured on one engine build. That does not extend to the two runtime-optimization fixtures of \Cref{sec:runtime-eval}, whose earlier records date from the v0.8.6 campaign, so both the hardware and the engine version changed between them. The relational searches of \Cref{sec:ec-eval} and \Cref{sec:maritime} are deterministic CPU procedures whose results do not depend on the host; the weight trainings beside them run on CPU as well. Rust crates are built in release mode; CUDA modules are staged cubin-first with PTX fallback and explicit JIT warm-up; Python components use \texttt{pyxlog} with CUDA-enabled PyTorch. The benchmark harness uses Criterion.rs with the settings in \Cref{tab:stat-protocol}, and GPU measurements include warm-up to amortize kernel-module loading and CUDA context setup.

Per-run measurement data is distributed with the paper for the head-to-head comparisons and overhead isolations, the circuit-cache ablation, the runtime-optimization ablations of \Cref{sec:runtime-eval}, and every rule-induction run of \Cref{sec:ecinduction,sec:maritime}. Each other figure below is stated together with the fixture, protocol and aggregation that produced it.

\begin{table}[htbp]
  \centering
  \caption{Benchmark harness settings.}
  \label{tab:stat-protocol}
  \small
  \begin{tabularx}{\linewidth}{@{}lL@{}}
    \toprule
    \textbf{Setting} & \textbf{Value} \\
    \midrule
    Sample size       & 10--100 runs (benchmark-dependent) \\
    Warm-up           & 3 iterations (PTX JIT, memory-pool init) \\
    Significance level & 0.10 \\
    Noise threshold   & 0.05 (ignore ${<}5\%$ variance) \\
    Seeding           & Deterministic LCG \\
    \bottomrule
  \end{tabularx}
\end{table}

\subsection{WCOJ vs.\ the binary-join plan}\label{sec:wcoj-eval}

This ablation asks how much the worst-case-optimal join path of \Cref{sec:datalog} saves over xlog's own binary-join plan on the skewed cyclic query the subsystem exists for. The query is a single three-way triangle join, the classic worst-case-optimal pathology; \Cref{tab:wcoj-protocol} gives the fixture and the timing protocol.

The keys are generated by four fixture variants named for the workloads they stand in for (call-graph edges, Andersen points-to~\cite{andersen1994}, \texttt{ddisasm}-style disassembly~\cite{ddisasm}, and a neural-symbolic mining analog) whose size parameters coincide at this scale, so the four cells are repeats of one workload and not four distinct ones. What the fixture does establish is skew: every join variable has a small hot domain, so the binary-join plan materializes an intermediate far larger than the final result, which is the regime WCOJ exists to address. WCOJ speedups depend strongly on input size and skew, and synthetic fixtures systematically overstate them, this one included, so the ratio below is the gain on the shape the subsystem targets, at this scale.

\begin{table}[htbp]
  \centering
  \caption{Triangle-join fixture and timing protocol for the WCOJ ablation.}
  \label{tab:wcoj-protocol}
  \small
  \begin{tabularx}{\linewidth}{@{}l L@{}}
    \toprule
    \textbf{Item} & \textbf{Value} \\
    \midrule
    Query              & three-way triangle join, one multiway join \\
    Key generation     & four fixture variants, coinciding size parameters \\
    Input              & three relations of $65{,}536$ rows each \\
    Output             & ${\approx}4.19$M triples \\
    Warm-up            & $20$ windows \\
    Reported time      & mean of $10$ timing windows \\
    Inner iterations   & $20$ per window (WCOJ), $40$ per window (binary join) \\
    Dispersion         & coefficient of variation reported, not gated on \\
    Correctness gate   & identical row sets on both paths before any timing \\
    Peak allocation    & about $2.3$\,GB, within the device's 12\,GB budget \\
    \bottomrule
  \end{tabularx}
\end{table}

\begin{figure}[htbp]
  \centering
  \includegraphics{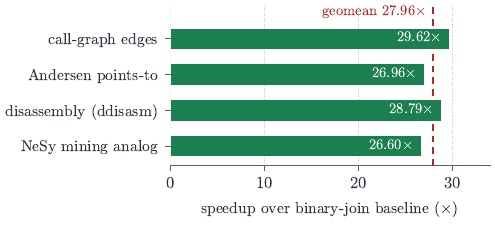}
  \caption{WCOJ vs.\ binary-join speedup on the four key-generation variants of the triangle fixture (\Cref{tab:wcoj-protocol}). Single-system ablation; the dashed line is the geometric mean.}
  \label{fig:wcoj}
\end{figure}

\Cref{fig:wcoj} reports the speedups: a $27.96\times$ geometric mean, with the four cells spanning $26.6\times$ to $29.6\times$. Since the variants coincide at this scale, that spread measures run-to-run variation and not robustness across distinct workloads. \Cref{sec:h2h} reports the external counterpart, a fused-counting comparison against Soufflé at millions of tuples.

On uniform or empty inputs the adaptive classifier selects the binary-join chain.

\subsection{Circuit caching and training}\label{sec:cache-eval}

This ablation asks how much of neural-symbolic training time circuit caching (\Cref{sec:prob}) removes by amortizing D4 compilation across epochs. Training is measured on MNIST single-digit addition; \Cref{tab:training-summary} carries both the full-run timing, in which the first epoch pays cold-start compilation and verification while warm epochs run only the cached forward/backward kernels, and the cache ablation beside it.

The ablation is a separate experiment from the full run: its uncached arm recompiles the circuit every epoch while its cached arm compiles once, so it isolates compilation amortization, and the apparent speedup therefore grows with epoch count toward the cold/warm ratio instead of being a fixed property. The 512-image protocol isolates cold/warm compilation timing and is deliberately under-trained (near-chance accuracy); held-out accuracy is measured at scale in the Scallop head-to-head (\Cref{sec:h2h}), where the same pipeline reaches ${\approx}95\%$.

\begin{table}[htbp]
  \centering
  \caption{MNIST-addition training summary. The timing rows come from a 512-image, 5-epoch training run at batch size 64; the ablation rows come from a separate experiment (512 images, 3 epochs, 3 seeds), whose speedup is the mean of its three per-seed ratios, not the ratio of its means.}
  \label{tab:training-summary}
  \small
  \begin{tabularx}{\linewidth}{@{}l l L@{}}
    \toprule
    \textbf{Metric} & \textbf{Value} & \textbf{Notes} \\
    \midrule
    First epoch (cold)         & $71.7$\,s  & D4 compile + CDCL verify \\
    Steady-state epoch (warm)  & $0.25$\,s  & cached forward/backward path \\
    Total training (5 epochs)  & $72.6$\,s  & \\
    Per-query cost (steady state) & $0.97$\,ms & forward + backward; steady epoch over 256 pairs, cold epoch excluded \\
    \addlinespace
    Ablation, uncached arm     & $242.9$\,s & mean training time, recompiling every epoch \\
    Ablation, cached arm       & $88.9$\,s  & mean training time, compiling once \\
    Circuit-cache speedup      & $2.74\times$ & 95\% CI $[2.29, 3.18]$ \\
    \bottomrule
  \end{tabularx}
\end{table}

\subsection{Head-to-head comparisons}\label{sec:h2h}

\Cref{tab:h2h} reports controlled comparisons against external engines, one per reasoning mode, on cloud GPUs (A100 80GB and L40S), with matched programs, data and metrics. Each artifact records the pod's CPU quota alongside its GPU, because that quota is what the CPU baselines are given: $26.35$ cores for Soufflé here, against $7.65$ in the previous measurement of the same comparison. xlog wins where its GPU-resident, bounded-memory design is the point, and does not where a mature CPU compiler is already fast enough on small inputs.

\begin{table*}[htbp]
  \centering
  \caption{Head-to-head comparisons: one external engine per reasoning mode, matched programs, data, metrics and seeds, on cloud GPU pods. The correctness and all-arm completeness gates pass for every comparison.}
  \label{tab:h2h}
  \small
  \begin{tabularx}{\linewidth}{@{}L l L L l@{}}
    \toprule
    \textbf{Mode / task} & \textbf{Baseline} & \textbf{xlog} & \textbf{Baseline} & \textbf{Ratio} \\
    \midrule
    Neural: MNIST addition; 20\,000 training images, 5 epochs, 3 seeds, held-out on the 10k test set
      & Scallop
      & acc $0.9561 \pm 0.0008$
      & acc $0.9468 \pm 0.0104$
      & accuracy parity; no timing claim \\
    \addlinespace
    Probabilistic: exact inference on five small programs
      & ProbLog2
      & $0$ error; per query flat at $0.79$--$0.87$\,s
      & $0$ error; per query $0.0085$--$0.0186$\,s
      & $45$--$100\times$ \emph{slower} \\
    \addlinespace
    Deterministic: triangle counting on hub-skewed graphs of $3.65$M to $17.84$M triangles
      & Soufflé (standalone compiled mode, 26 jobs)
      & counts match at every size; $0.57$--$0.68$\,s, nearly flat in the input; peak device allocations $85$--$1{,}033$\,MB
      & $0.51$--$3.77$\,s, growing with the input; one-time per-case compilation recorded separately
      & Soufflé/xlog: $0.88\times \to \mathbf{5.54\times}$ \\
    \bottomrule
  \end{tabularx}
\end{table*}

\textbf{Neural-symbolic: MNIST addition vs.\ Scallop.} With identical MNISTNet, data, metric and seeds, xlog and Scallop reach comparable held-out accuracy: $0.9561 \pm 0.0008$ against $0.9468 \pm 0.0104$ on the 20k protocol, and both near chance on the under-trained 512-image protocol. The two systems reach that accuracy by different routes: Scallop scores each fact through a bounded set of proof clauses, while xlog evaluates the exact weighted model count of the compiled circuit on every step. \emph{We make no per-epoch speed claim on this comparison.} Scallop's steady epoch is not monotone in the pod's CPU quota --- $31.79$\,s at $13.6$ cores and $54.99$\,s at $27.2$ on the same card, the same wheel and the same seeds --- so an xlog/Scallop epoch ratio measured on any one host reports that host as much as either engine. Establishing the ratio would take that curve measured across quotas; the accuracies above stand independently of it.

\textbf{Probabilistic: exact inference vs.\ ProbLog2.} On five programs (a conditioned Bayesian fragment and probabilistic reach chains) xlog's exact query probabilities match the analytic answers to zero error, identical to ProbLog2 (correctness gate $|\Delta| < 10^{-4}$). On timing, however, xlog is \emph{not} competitive on these small programs. Its per-query time is flat, a fixed GPU launch plus a per-call D4 compile and CDCL verify, while ProbLog2's mature CPU compiler runs each query $45$--$100\times$ faster. We therefore make no exact-inference \emph{speed} claim; the probabilistic contribution is correctness-equivalent inference with a CUDA-resident circuit data plane and certification of the final smoothed circuit. The host orchestrates the route and observes bounded solver control state, and its fixed costs amortize only at scale or under repeated evaluation.

\textbf{Deterministic: WCOJ triangle counting vs.\ Soufflé.} On hub-skewed graphs, xlog's fused worst-case-optimal triangle counting, xlog's enumerate-then-count diagnostic, and Soufflé produce the same per-root triangle totals at every size. The comparison uses the median of three full-process executions after one-time native builds: a standalone Soufflé executable is generated once per case with twenty-six jobs, matching the pod's CPU quota, and its $13.05$--$13.56$\,s compilation is recorded separately. \emph{The defensible claim is how the ratio moves with input size.} Relative to fused xlog execution, compiled Soufflé is $0.88\times$, $1.96\times$, $2.59\times$, $3.88\times$ and $5.54\times$ across $150$k to $1.2$M edges: on the smallest graph Soufflé is the faster of the two, and xlog only overtakes it from $300$k edges on. What holds monotonically is that the ratio grows with the input, because xlog's execution time is nearly flat ($0.57$--$0.68$\,s) while Soufflé's grows linearly ($0.51$--$3.77$\,s). That flatness needs a caveat the artifact supplies: the engine's own counting time is $25$--$43$\,ms of it, so between $93$ and $96\%$ of the xlog wall time here is process start, CUDA context creation and Arrow input. What is flat at these sizes is the fixed cost, and the engine time is small enough that this comparison does not resolve how it scales. Fused counting never materializes the triangles and its peak device allocations are $85$ to $1{,}033$\,MB, while enumerate-then-count peaks at $3{,}287$ to $44{,}979$\,MB. All arms complete, both cross-system count checks pass, and the all-arm comparison gate is therefore true. This supports bounded intermediates on these skewed, large-intermediate workloads, not a universal Datalog speed claim: native compilation time is excluded from the execution medians, the measurements are hardware-specific and quota-specific, and on moderate skew the advantage over xlog's own binary join is $2.00$--$4.95\times$ --- measured against a binary arm that completes at every size there, where the earlier artifact had recorded its two largest cases as skipped for blow-up.

\subsection{Event-Calculus rule induction}\label{sec:ec-eval}

This subsection asks how the induced theories score against the figures other authors have published for the same benchmark, and what the runs that do not favour the system look like. \Cref{sec:ecinduction} states the induction protocol and analyses its runs, and \Cref{tab:ecinduction} there lists them; \Cref{tab:ec-runs} collects their frame-level scores.

\textbf{Published figures for the same benchmark.} \Cref{tab:ec-published} places our pre-registered ten-fold row beside the four published frame-level \texttt{holdsAt} F1 figures for \texttt{meeting} on the whole distributed CAVIAR corpus. Our figure is \emph{below the floor of the published span}, by $0.0021$: a fully pre-registered run lands just under the lowest published figure, which is the hand-crafted rule set. The second-iteration run reaches $0.7782$, which is inside the span, but it is one disclosed adaptive iteration (\Cref{sec:ecinduction}); it is not the row in this table.

\textbf{The direct-protocol reference on the same folds.} The run that produced that row also scores a direct per-timestep reference theory on the same ten folds, under the direct protocol's own pre-registered defaults: held-out accuracy, a fixed $0.75$ fit gate, the default tie floor. It lands far below the Event-Calculus row beside it (\Cref{tab:ec-runs}). The same reference theory comes out of all three ten-fold runs (the pre-registered one, the second iteration and the neural negative), so the second iteration's added vocabulary never reaches the direct search. Its recall is dominated by one fold on which the direct search abstains outright, leaving 855 gold frames undetected. The figure is what those defaults produce on this corpus; it is no ceiling for the direct protocol, whose studied settings produce the windowed-fold results of \Cref{sec:ecinduction}.

\begin{table*}[htbp]
  \centering
  \caption{Frame-level \texttt{holdsAt} F1 for the \texttt{meeting} fluent on the distributed CAVIAR corpus, micro-averaged over pooled counts. Rows are ordered alphabetically by system name, not by F1: the five protocol deltas printed beneath the table do not all act in the same direction. Dashes are figures the cited table does not report.}
  \label{tab:ec-published}
  \small
  \begin{tabularx}{\linewidth}{@{}l c c c L@{}}
    \toprule
    \textbf{System} & \textbf{P} & \textbf{R} & \textbf{F1} & \textbf{Source} \\
    \midrule
    Hand-crafted rules                      & $0.644$  & $0.855$  & $0.735$  & OLED~\cite{oled} Tab.~1(b), \texttt{EC\_crisp}; the same value appears in WOLED~\cite{woled} Tab.~2 \\
    OLED                                    & $0.678$  & $0.953$  & $0.792$  & OLED~\cite{oled} Tab.~1(b), whole dataset \\
    OLED, as re-run by other authors        & ---      & ---      & $0.782$  & WOLED~\cite{woled} Tab.~2 \\
    This work, EC${}+{}$inertia, pre-registered & $0.6580$ & $0.8271$ & $0.7329$ & 10-fold CV by video segment, all gates re-derived per fold (\Cref{sec:ecinduction}) \\
    WOLED-ASP                               & ---      & ---      & $0.887$  & WOLED~\cite{woled} Tab.~2 \\
    \bottomrule
  \end{tabularx}

  \vspace{0.5em}
  {\footnotesize\raggedright
  \emph{Protocol deltas, which must travel with this table.}
  (i)~The published runs cross-validate over their own windowed interpretations with subsampled negatives; ours folds over whole video segments.
  (ii)~They learn full initiation \emph{and} termination programs; our row commits a two-literal initiation clause and an empty termination theory. Out of its 2{,}304 predicted-positive frames, 781 of the 788 false positives fall on the folds with detected interior terminations; the other 7 are initiation-side errors persisted by inertia on a fold whose test side detects none.
  (iii)~Their rule language admits longer bodies---bottom clauses averaging some fifteen literals against our three-literal cap.
  (iv)~Their settings are reported as the best among several tried; ours were pre-registered. No number of ours is adjusted on that basis.
  (v)~All rows compute F1 on \texttt{holdsAt} inferred through inertia, on the same corpus, the one axis on which the protocols do match, and the reason the rows are placed side by side at all.
  \par}
\end{table*}

The remaining runs of \Cref{sec:ecinduction}, including the two that do not favour the system, are scored in \Cref{tab:ec-runs} and diagnosed there. One of those rows needs a disclosure that section does not carry.

\textbf{The \texttt{moving} threshold, disclosed.} The two fluents carry different canonical proximity thresholds in the published CAVIAR rule set (\texttt{meeting} over \texttt{close\_25}, \texttt{moving} over \texttt{close\_34}), and the \texttt{moving} figure above is measured under the canonical \texttt{close\_34}. An earlier run over the same folds measured \texttt{moving} under the \texttt{meeting} threshold of 25 and scored \emph{higher} (\Cref{tab:ec-runs}). The canonical threshold remains the right headline, and the higher number is not evidence against it: where the clause is selected, widening \texttt{close} from 25 to 34 behaves exactly as threshold semantics predict, recall up and precision down.

The micro figure falls for a selection-level reason instead. On the fold carrying 1{,}546 of the 3{,}136 gold \texttt{moving} frames, the wider candidate pool puts the top two candidates inside the selection tie band, at a margin of $0.0093$ against the $0.01$ tie floor, and the search abstains there where the threshold-25 run had committed. That figure therefore leaned on a fold whose selection was tie-fragile, which is a reason to report the lower canonical number and the disagreement together rather than either alone.

\textbf{The maritime figures are not tabulated beside the published ones.} The maritime runs of \Cref{sec:maritime} are not placed beside the published figure for that fluent; \Cref{subsec:maritime-comparison} says why.

\begin{table*}[htbp]
  \centering
  \caption{Frame-level \texttt{holdsAt} scores for every rule-induction run of \Cref{sec:ecinduction}, micro F1 over pooled per-fold counts. The distributed corpus is the dump shipped with the published Event-Calculus learners and carries the duplication defect measured in \Cref{sec:ecinduction}; the clean corpus is its deduplicated, scene-family-grouped form. Dashes are quantities the run does not report separately.}
  \label{tab:ec-runs}
  \small
  \begin{tabularx}{\linewidth}{@{}L l c c c l@{}}
    \toprule
    \textbf{Run} & \textbf{Corpus} & \textbf{P} & \textbf{R} & \textbf{F1} & \textbf{tp\,/\,fp\,/\,fn} \\
    \midrule
    \texttt{meeting}, Event-Calculus route, pre-registered      & distributed & $0.6580$ & $0.8271$ & $0.7329$ & $1516$\,/\,$788$\,/\,$317$ \\
    \texttt{meeting}, direct-protocol reference, same folds     & distributed & $0.6864$ & $0.1266$ & $0.2137$ & $232$\,/\,$106$\,/\,$1{,}601$ \\
    \texttt{meeting}, Event-Calculus route with learned \texttt{close\_nn} & distributed & $0.1250$ & $0.0005$ & $0.0011$ & $1$\,/\,$7$\,/\,$1{,}832$ \\
    \addlinespace
    \texttt{meeting}, Event-Calculus route                      & clean & --- & --- & $0.0$ & $0$\,/\,$120$\,/\,$1812$ \\
    \texttt{meeting}, direct route                              & clean & --- & --- & $0.0$ & $0$\,/\,$106$\,/\,$1812$ \\
    \texttt{moving}, Event-Calculus route, no initiation clause on any fold & clean & --- & --- & --- & $0$\,/\,$0$\,/\,$3136$ \\
    \texttt{moving}, direct route, canonical \texttt{close\_34} & clean & $0.5334$ & $0.3817$ & $0.4450$ & $1197$\,/\,$1047$\,/\,$1939$ \\
    \texttt{moving}, direct route, earlier run under \texttt{close\_25} & clean & --- & --- & $0.4868$ & $1295$\,/\,$890$\,/\,$1841$ \\
    \bottomrule
  \end{tabularx}
\end{table*}

\subsection{Verification cost and residency benefit}\label{sec:overhead}

These two isolations ask what xlog's design costs, separated from ordinary GPU acceleration: what the equivalence proof buys with cold-compile time, and what device residency saves in transfers. Both run on an A100 80GB PCIe pod with a $26.35$-core CPU quota, and \Cref{tab:overhead} carries their measurements.

\textbf{Verification cost.} The cold epoch's compilation time is dominated by the on-GPU CDCL equivalence check rather than the D4 compile. Split on probabilistic reachability chains, CDCL verification accounts for $98.8$--$99.4\%$ of the cold-compile time and grows with circuit size, while D4 compilation stays between $1.2$ and $11.7$\,ms and is non-zero at every point. The ``expensive'' cold epoch is thus almost entirely the price of the machine-checked correctness certificate (\Cref{sec:prob}), a cost the CPU-symbolic and GPU-Datalog systems surveyed in \Cref{sec:related} do not pay, because they do not verify.

\begin{table}[htbp]
  \centering
  \caption{Design-cost isolations. Verification splits the cold compile on probabilistic reachability chains of length $n$. Residency forces a device$\to$host$\to$device round-trip at the neural--symbolic boundary, swept over $2$ to $512$ handoffs per step on the batched addition path.}
  \label{tab:overhead}
  \small
  \begin{tabularx}{\linewidth}{@{}l L@{}}
    \toprule
    \textbf{Setting} & \textbf{Measured} \\
    \midrule
    \multicolumn{2}{@{}l}{\emph{Verification cost}} \\
    chain $n{=}5$        & CDCL verify $531$\,ms \\
    chain $n{=}40$       & CDCL verify $2{,}633$\,ms \\
    \addlinespace
    \multicolumn{2}{@{}l}{\emph{Residency benefit}} \\
    per handoff          & round-trip of roughly $61$--$176\,{\mu}$s \\
    $32$ handoffs/step   & $9.2\%$ of the step \\
    $128$ handoffs/step  & $16.1\%$ of the step; this is the standard batch-$64$ MNIST-addition step, $9.6$\,ms of $59.3$\,ms \\
    $512$ handoffs/step  & holds near $15.5\%$ of the step \\
    \bottomrule
  \end{tabularx}
\end{table}

\textbf{Residency benefit.} To isolate the transfer-free advantage from ordinary GPU acceleration we run the same neural--symbolic pipeline with and without a forced device$\to$host$\to$device round-trip at the neural--symbolic boundary, the transfer a CPU-reasoning hybrid pays every step, sweeping the number of neural$\to$symbolic handoffs per step. Below a handful of handoffs the round-trip cost is lost in run-to-run variation. From a few dozen upward it is a visible share of the step, because the batched circuit compute grows sublinearly with the batch while the transfer grows linearly with the handoff count.

On a single query the transfer is therefore negligible ($2.7\%$ at two handoffs); at a realistic training batch it is a ${\approx}16\%$ tax that xlog does not pay, distinct from and additive to the GPU-compute and circuit-caching advantages. The forced per-buffer copy is an upper bound, since a hybrid could coalesce transfers. The larger transfer-dominated regime, large-state Datalog fixpoints where whole relations would cross the bus every iteration (\Cref{sec:intro}), lies outside what this isolation measures (\Cref{sec:limits}).

\subsection{Index reuse and the chain scorer}\label{sec:runtime-eval}

This pair of ablations asks what two of the runtime optimizations of \Cref{sec:datalog} save on the repeated-session workloads they target, each against the path it replaces. The persistent hash-index manager retains built join indices across evaluations, keyed on relation generation, schema signature and device ordinal, under a byte-budgeted LRU policy: a session whose relations change little between calls reuses an index instead of rebuilding it, and a generation change invalidates the entry instead of serving a stale one. The profile-gated shared-memory chain scorer tiles the left relation through shared memory when scoring chain-shaped candidate bodies during exact rule induction, and is gated on a candidate-row threshold below which the baseline scorer runs unchanged.

Both target the same waste, work repeated across calls that the previous call already did, and both are transparent to the program. Common-subexpression elimination and adaptive re-optimization are held to output equivalence with the unoptimized plan.

\Cref{tab:runtime-opt} gives the medians. Two things belong with them. The index fixture was re-measured in release after an earlier version of its artifact recorded a build flag the run had not used; the same pod then ran both profiles back to back, and release is the \emph{higher} ratio, $8.33\times$ against $7.72\times$ in debug. The reason is that the two arms sit on different sides of the bus: the index-rebuilding arm is GPU-bound and barely moves between profiles ($15.83$ to $15.78$\,ms), while the cached arm is host-side and does get faster ($2.05$ to $1.89$\,ms), so optimizing the build widens the gap. And the chain scorer's ratio rose from $5.58\times$ to $7.20\times$ partly because its baseline arm got slower on this card, $27.51$ to $52.42$\,ms, so that gain is not a like-for-like improvement over the earlier record. The index fixture is build-heavy by construction, so that building the index dominates the join, and neither arm records a host transfer. The chain scorer's two arms do not overlap on the chain-hot fixture; on a small input below its gate threshold, where the baseline path runs instead, the two medians differ by $2.0\%$ and the arms' ranges overlap. Both are point medians on single synthetic fixtures, and the scorer's coverage output is identical to the baseline's on both. The scorer's artifact now also records the host-transfer count per arm rather than asserting a fixed one: what the claim rests on is that the shared-memory arm adds none relative to baseline, and the measured difference is zero.

\begin{table}[htbp]
  \centering
  \caption{Runtime-optimization ablations, each a median over the timed runs of one synthetic fixture.}
  \label{tab:runtime-opt}
  \small
  \begin{tabularx}{\linewidth}{@{}L L@{}}
    \toprule
    \textbf{Fixture} & \textbf{Baseline median $\to$ optimized median} \\
    \midrule
    Persistent index reuse: repeated-session semi-join, eight left rows probing eight million right rows
      & $15.78$\,ms $\to$ $1.89$\,ms, a $8.33\times$ gain over $9$ timed runs after $12$ warm-up runs \\
    \addlinespace
    Shared-memory chain scorer: chain-hot body search, $768$ candidate rows against a gate threshold of $256$
      & $52.42$\,ms ($52.41$--$52.97$) $\to$ $7.28$\,ms ($7.28$--$7.29$), a $7.20\times$ gain over $12$ timed runs after $3$ warm-up runs \\
    \bottomrule
  \end{tabularx}
\end{table}

\subsection{Capability coverage across systems}

\Cref{tab:comparison} positions xlog qualitatively against representative systems; each addresses a subset of the design space, and xlog's contribution is exposing the supported routes through one typed language and provider-owned CUDA runtime services. This table covers capability: the quantitative head-to-head results against Scallop, ProbLog2 and Soufflé are in \Cref{sec:h2h}.

The rule-induction row covers two paths of different maturity: the trainable-rule-body path is carried onto an external benchmark in \Cref{sec:ecinduction} and tabulated in \Cref{sec:ec-eval}, while the differentiable-ILP path is not exercised there at all (\Cref{sec:nesy}). The table's ``yes'' claims capability, not maturity (\Cref{sec:limits}).

\begin{table*}[htbp]
  \centering
  \caption{Capability comparison across neurosymbolic and GPU logic systems.}
  \label{tab:comparison}
  \small
  \begin{tabularx}{\linewidth}{@{}l L L L L L@{}}
    \toprule
    \textbf{Dimension} & \textbf{DeepProbLog} & \textbf{Scallop} & \textbf{Lobster} & \textbf{GPUlog} & \textbf{xlog} \\
    \midrule
    Symbolic execution      & CPU (Prolog) & CPU (Datalog)    & GPU (Datalog)      & GPU (Datalog) & GPU (semi-naive) \\
    Probabilistic inference & Exact (CPU)  & Provenance (CPU) & Provenance (GPU)   & None       & Exact d-DNNF + MC (GPU) \\
    Circuit verification    & None         & None             & None               & ---        & On-GPU CDCL \\
    Neural integration      & Prolog bridge & Differentiable  & Differentiable (GPU) & None     & Zero-copy, fused \\
    Zero-copy ML interop    & No           & Partial          & Partial            & No         & Yes \\
    Rule structure induction & No          & No               & No                 & No         & Yes (beta): trainable bodies, dILP \\
    Epistemic reasoning     & No           & No               & No                 & No         & Yes (finite fragment) \\
    \bottomrule
  \end{tabularx}
\end{table*}

\subsection{Log-evidence, rule unions, and exact constraint solving}\label{sec:noquant}

The capabilities below round out the runtime surface. Each is described for what it does; no timing is claimed for any of them.

\textbf{Exact log-evidence and the CNF-variable-to-fact map.} An exact evaluation now returns the log partition function of the weighted formula, $\log Z$, alongside the per-query probabilities, so a program's evidence mass is read off directly instead of being reconstructed from marginals; it surfaces as \texttt{log\_z\_e} on the Python result. Paired with it, \texttt{prob\_var\_map()} reports which probabilistic fact each CNF variable stands for (an ordinary fact, one Bernoulli decision of an annotated disjunction's chain, or padding) so the per-variable gradient buffers can be attributed back to program facts and not to opaque indices. Its length is the encoder's variable \emph{capacity}, not a variable count, and it is refused with a typed error where there is no CNF encoding to report: Monte Carlo programs, and exact programs taking the GPU fast path that skips the encoding.

\textbf{Same-head rule unions in one multiway pass.} A predicate head defined by many rules used to fold its per-rule contributions into the head relation one union at a time, re-sorting the growing accumulator once per rule and going quadratic in the rule count, a real cost at corpus scale where one head can carry thousands of rules. Contributions are now merged with a single multiway union per head per iteration. The union count therefore no longer scales with the same-head rule count (eight same-head rules record one union, not eight) while the derived rows stay byte-identical to the unbatched result.

\textbf{An exact stage beyond enumeration capacity.} Constraint solving over several interacting constraints enumerates a component completely where it is small enough. Long chain-shaped components that exceed that capacity take a memoized dynamic-programming stage on the GPU instead, which computes the same totals exactly instead of approximating them. A component wider than the configured limit is refused with a typed error instead of served by an approximation, and the solver's buffers reach Python through the same zero-copy DLPack~\cite{dlpack} views as the rest of the runtime (\Cref{subsec:interop}).

\section{Related Work}\label{sec:related}

xlog draws on five lines of research: neurosymbolic programming, GPU-accelerated Datalog, probabilistic logic programming, GPU SAT, and differentiable ILP. Its contribution is to expose those capabilities through one typed language and shared CUDA services while retaining route-specific execution and differentiability contracts.

\subsection{Neurosymbolic programming}

DeepProbLog~\cite{deepproblog} replaces selected probabilistic facts with neural-network outputs, enabling end-to-end gradient training of neural predicates inside a probabilistic-logic framework. NeurASP~\cite{neurasp} integrates network outputs as probability annotations on answer-set programs. Both perform symbolic inference on the CPU and shuttle gradients across the bus. Scallop~\cite{scallop} provides a differentiable Datalog with provenance-semiring reasoning~\cite{green2007provenance} and a relational symbolic representation, and is the closest language-level analogue to xlog. Its reasoning core, however, is CPU-based.

Lobster~\cite{lobster} is the most directly comparable system: it GPU-accelerates neurosymbolic programs in the Scallop style, and shares xlog's premise that the symbolic data plane belongs on the GPU. Against it, xlog is complementary along three axes. It includes exact knowledge compilation and epistemic reasoning in addition to provenance-based differentiable Datalog. It certifies the final smoothed circuit's logical equivalence to its back-end source formula using GPU CDCL and on-device proof checking. And it reports route-specific residency with byte-level transfer accounting. No head-to-head performance comparison with Lobster is drawn here (\Cref{sec:limits}).

Differentiable-SAT and logic-as-loss methods (SATNet~\cite{satnet}, semantic loss~\cite{semanticloss}, and Logic Tensor Networks~\cite{ltn}) instead couple neural models to constraints through relaxed or fuzzy surrogates. By contrast, xlog couples through \emph{exact} weighted model counting, trading their broader applicability for exact gradients and the on-device verifiability of \Cref{sec:prob}.

\subsection{GPU-accelerated Datalog and joins}

GPUlog~\cite{gpulog} ported bottom-up fixed-point computation to CUDA using HISA indexing, and reported up to $45\times$ speedups over CPU Datalog engines. VFLog~\cite{vflog} advanced this with a vertically-fused, column-oriented kernel design that avoids intermediate materialization, reporting up to $200\times$ over CPU column engines; mnmgDatalog~\cite{mnmgdatalog} extends GPU Datalog to multi-node, multi-GPU clusters. Most relevant to xlog's join path, Sun et al.~\cite{sungpuwcoj} scale worst-case-optimal joins to GPUs.

All of them target deterministic Datalog only: none supports probabilistic semantics, weighted model counting, or gradient computation over derived facts. Like them, xlog uses a columnar, kernel-fused execution model, and it also promotes eligible cyclic and multiway rules to a worst-case-optimal join path~\cite{ngo2018wcoj,veldhuizen2014leapfrog,wang2023freejoin}. But that engine is one component of a substrate that extends through circuit-based inference to neural-symbolic gradient propagation in a single address space.

\subsection{Probabilistic logic programming}

ProbLog2~\cite{problog2} established the modern pipeline for exact probabilistic inference: ground the program, encode provenance as a propositional formula, compile to a tractable target (d-DNNF~\cite{darwiche2001dnnf} or SDD~\cite{sdd}), and evaluate weighted model counts~\cite{chaviraDarwiche2008}. The compilers (D4~\cite{d4compiler}, c2d~\cite{c2d}) run as host-side subprocesses, and recent work makes such compilation \emph{certified}~\cite{capelli2019}. xlog adopts the same compile-once/evaluate-many model with CUDA-backed data planes for provenance, Tseitin encoding, D4 compilation, equivalence verification, and forward/backward evaluation. The host orchestrates those stages and reads bounded solver control state; device circuit and gradient buffers share an address space with the neural and deterministic data planes.

\subsection{GPU SAT}

ParaFROST~\cite{parafrost} parallelizes CDCL clause-database simplification on the GPU while retaining sequential conflict-driven search on the CPU. FastFourierSAT~\cite{fastfouriersat} reformulates SAT as continuous optimization and applies GPU local search, an incomplete solver for satisfiable instances. Both are standalone competition solvers. In xlog the GPU CDCL module is instead a verification component inside knowledge compilation (\Cref{sec:prob}), proving circuit--formula equivalence via two UNSAT proofs and providing the certificates the probabilistic and epistemic backends rely on.

\subsection{Differentiable ILP}

Differentiable ILP~\cite{dilp} casts first-order rule induction as differentiable optimization, scoring candidate rules by soft forward-chaining. Dense materialization over the rule space scales cubically in the number of constants, and implementations remain CPU-based. The dILP subsystem of xlog (\Cref{sec:nesy}) replaces dense materialization with sparse GPU bitmasks and on-device scatter--gather credit assignment, avoiding the cubic blowup while preserving differentiability.

\subsection{Logic programming languages}

xlog inherits ideas from a long line of logic languages. Prolog~\cite{swiprolog} offers unification and metaprogramming on a CPU interpreter; Mercury~\cite{mercury} introduced strong typing and modes, compiling to native code. Souffl{\'e}~\cite{souffle} is a typed Datalog compiling to C++ for program analysis, and clingo~\cite{clingo} provides answer-set semantics on CPU grounders and solvers. From these xlog adopts the typed-predicate discipline while unifying typed Datalog, probabilistic facts, neural predicates, and epistemic operators in one language whose supported routes use CUDA kernels and provider-owned runtime services.

\section{Limitations and Future Work}\label{sec:limits}

\subsection{Current limitations}

\textbf{NVIDIA GPU required.} The execution kernels are written in CUDA C; there is no OpenCL, Metal, or ROCm backend. The persistent-index manager, recorded-launch discipline, and in-kernel hash-consing use CUDA-specific primitives such as cooperative groups, warp-level operations, and unified virtual addressing. Users without a supported NVIDIA GPU cannot run xlog.

\textbf{Device memory and fixed capacities are bounded.} Relations, indices, circuit buffers, and Monte Carlo sample arrays are allocated on-device; there is no out-of-core spilling. An allocation that exceeds the configured byte budget fails with \texttt{RESOURCE\_EXHAUSTED}. A row, tuple, sample, domain, or other fixed-count limit instead fails with \texttt{CAPACITY\_EXCEEDED}, naming the exhausted unit. A bounded recursive or Monte Carlo fixpoint that reaches its iteration limit without convergence fails with \texttt{CONVERGENCE\_FAILURE}; partial counters do not turn it into a result.

\textbf{Maturity of the rule-learning surfaces.} The differentiable-ILP subsystem (\Cref{sec:nesy}) is beta: the search space is restricted to definite programs, convergence is sensitive to learning-rate schedules, and its API may change. The trainable-rule-body surface of \Cref{subsec:trainable-bodies} (mixed neural/symbolic bodies, existential joins, and joint multi-rule mixtures) is at the same maturity. It is carried onto a public benchmark under pre-registered gates and cross-validation in \Cref{sec:ecinduction}, but how its accuracy and training cost compare with dedicated relational learners on standard benchmarks is an open question.

\textbf{Coverage of the epistemic runtime.} The epistemic runtime (\Cref{sec:epistemic}) executes its accepted data-plane work on the GPU, but the broader solver portfolio (MaxSAT/portfolio services) and the end-to-end probabilistic--epistemic evidence path are still being broadened. The negated-modal WFS route and Gelfond-1991 compatibility refinement are GPU-backed but host-orchestrated. Positive modal dependency cycles have defined execution: FAEEL computes a founded least fixpoint, including an empty extension for an unseeded cycle, and supported Gelfond-1991 positive \texttt{possible} cycles compute the greatest compatible set of concrete tuples from an upper bound and frozen snapshots. The shared recursion-depth setting bounds both descending compatibility refinements and WFS iterations.

\textbf{Fail-closed boundaries.} Raw lowering of modal literals, recursive epistemic programs that also contain modal integrity constraints, WFS shapes outside the supported negated-modal plan, unbounded tuple-key forms, and recursive negation or aggregation within a Gelfond-1991 compatibility component all remain fail-closed. The Python batch-query helpers are typed but do not yet accept floating-point uploads, a convenience-API gap and not a core execution limit.

\textbf{Partial head-to-head coverage.} \Cref{sec:h2h} reports controlled comparisons on identical programs and data against one external engine per reasoning mode: Scallop (neural-symbolic), ProbLog2 (probabilistic), and Soufflé (deterministic). Coverage is still partial. Lobster, the most directly comparable GPU-neurosymbolic system, and the GPU Datalog engines GPUlog and VFLog are not among the baselines; DeepProbLog was not run under a matched (non-timeout) protocol; and each comparison rests on a single GPU configuration rather than a cross-hardware sweep. That last limitation is not hypothetical: measuring the MNIST comparison on two L40S pods with the same wheel, the same seeds and different CPU quotas moved Scallop's steady epoch from $31.79$\,s at $13.6$ cores to $54.99$\,s at $27.2$, so the epoch ratio between the two systems changed by a factor of $1.8$ with neither engine touched. The pods differed in driver revision and host as well as in quota, so the quota is the candidate explanation rather than the isolated cause; the $27.2$-core measurement is distributed as the companion file beside the MNIST artifact. That is why \Cref{sec:h2h} reports accuracy for that comparison and no timing. A per-cell ratio measured on one host is a property of the pair \emph{and} the host; broader multi-system, multi-hardware comparison remains future work.

\textbf{Rule-induction evidence.} The evidence of \Cref{sec:ecinduction} is thinnest exactly where it is leak-free: the deduplicated, scene-family-grouped corpus carries too few initiation events for a cross-validated claim. Both routes commit a clause on most folds and that clause does not transfer, so the protocol yields no \texttt{meeting} theory and supports no conclusion about that fluent's learnability in either direction. The protocol-matched numbers are computed instead on the dump distributed with the published Event-Calculus learners, which carries a measured duplication defect; we report there because it is the regime the published figures share, and every figure computed on it inherits that caveat. No ordering against OLED, WOLED-ASP or the hand-crafted rule set follows from \Cref{tab:ec-published}, and none is claimed: the five protocol deltas printed with that table are real, unquantified, and do not act in the same direction, while the clean protocol has too few events to settle any of them. Learning termination programs and lifting the body-length cap, which would remove two of those deltas, is future work.

\textbf{Two figures of different status.} The headline figure of \Cref{sec:ecinduction} is fully pre-registered: gates, seeds and folds were fixed before the run, and that run was executed once. The second-iteration figure differs from it in exactly one respect, its Event-Calculus vocabulary having been chosen after the first run's failure mode was known on the same folds, which makes it one disclosed adaptive iteration and not a pre-registered result; it is reported only with that label and is never the row we place beside a published number.

\textbf{Maritime scope.} The maritime runs of \Cref{sec:maritime} remove the evidence-starvation bound, but they carry bounds of their own. Their supervision is synthetic: the gold labels are a hand-crafted rule's output over the same streams the vocabulary is derived from, a regime the published maritime figures share. Every number there is therefore a rule-reconstruction figure under an incomplete vocabulary, not a detection figure, and none of those runs exercises the GPU substrate. Their temporal grid is the critical-point grid of the RTEC exports, not the dense grid of the published figure, so no maritime number of ours stands beside the published $0.98$ and no pass time beside the published training times (\Cref{subsec:maritime-comparison}). And the single-pass runs are online in the weights only, their body pool and permutation-null gate being computed on the training side: they demonstrate single-pass weight learning and not online structure revision, which remains future work, and the zero degradation they measure rests on a single body sitting above the decision threshold in every regime, on a $0.007$ margin; it is a property of this pool, corpus and threshold, not a robustness result.

\subsection{Future work}

\textbf{Out-of-core execution.} A streaming mode would partition relations across host and device, paging tiles through GPU memory in fixpoint-iteration order, removing the single-GPU-memory ceiling at the cost of added transfers.

\textbf{Multi-GPU partitioned evaluation.} Inspired by mnmgDatalog's~\cite{mnmgdatalog} radix-hash partitioning and GPU-aware all-to-all shuffle, a multi-GPU backend would distribute relations across devices and coordinate joins over NVLink or PCIe.

\textbf{Deeper epistemic residency.} The remaining epistemic work is stronger residency and coverage: a device-resident, no-host-interaction well-founded path, broader non-finite tuple-key forms, and semantic forms beyond the current finite accepted surface.

\section{Conclusion}\label{sec:conclusion}

xlog is a CUDA-native logic programming engine with one typed language and
provider-owned runtime services for deterministic, probabilistic, epistemic,
and neural-symbolic use. A neural network is an ordinary predicate; exact query
gradients flow through the probabilistic knowledge-compilation route, while the
other routes reuse typed device-buffer interfaces without claiming identical
differentiability. The final smoothed probabilistic circuit is certified against
its back-end source formula before it is cached or evaluated. Solver proof traces
are checked on the GPU, and the host observes bounded status and error scalars.

Residency claims are likewise route-specific. Ordinary relational and exact
execution are host-orchestrated over device data planes. Eligible recursion and
resident Monte Carlo use certified cores that record zero tracked transfers and
then return one authoritative terminal receipt. Byte exhaustion, fixed-capacity
exhaustion, and nonconvergence remain separate typed outcomes.

The measurements describe a system that pays off where the symbolic workload
is large and repeated: reusing a verified circuit across training iterations,
and avoiding the intermediate blowup of binary-join plans on skewed cyclic
queries. Against external engines the picture is mixed: clear
gains in bounded-memory join evaluation and in per-epoch neurosymbolic
training, and a clear loss to a mature CPU compiler on small exact-inference
programs.

What they do not give is a general ranking of xlog against other systems. Each
comparison covers one engine per reasoning mode on one class of program, the
rule-induction results rest on two public corpora, and the epistemic and
rule-learning surfaces are young. The evidence supports the architecture, not
a claim of universal superiority.

The direction is the one the architecture already implies: out-of-core and
multi-GPU execution to lift the memory ceiling, deeper residency for the
epistemic routes, and structure learning that runs on the stream rather than
over it.

\printbibliography

@inproceedings{gpulog,
  author    = {Sun, Yihao and Shovon, Ahmedur Rahman and Gilray, Thomas and Kumar, Sidharth and Micinski, Kristopher},
  title     = {Optimizing {D}atalog for the {GPU}},
  booktitle = {Proceedings of the 30th ACM International Conference on Architectural Support for Programming Languages and Operating Systems (ASPLOS)},
  year      = {2025},
  doi       = {10.1145/3669940.3707274},
}

@inproceedings{vflog,
  author    = {Sun, Yihao and Kumar, Sidharth and Gilray, Thomas and Micinski, Kristopher},
  title     = {Column-Oriented {D}atalog on the {GPU}},
  booktitle = {Proceedings of the AAAI Conference on Artificial Intelligence},
  volume    = {39},
  number    = {14},
  pages     = {15177--15185},
  year      = {2025},
  doi       = {10.1609/aaai.v39i14.33665},
}

@inproceedings{mnmgdatalog,
  author    = {Shovon, Ahmedur Rahman and Sun, Yihao and Gilray, Thomas and Micinski, Kristopher and Kumar, Sidharth},
  title     = {Multi-Node Multi-{GPU} {D}atalog},
  booktitle = {Proceedings of the 39th ACM International Conference on Supercomputing (ICS)},
  year      = {2025},
  doi       = {10.1145/3721145.3730431},
}

@article{problog2,
  author  = {Fierens, Daan and {Van den Broeck}, Guy and Renkens, Joris and Shterionov, Dimitar and Gutmann, Bernd and Thon, Ingo and Janssens, Gerda and {De Raedt}, Luc},
  title   = {Inference and Learning in Probabilistic Logic Programs using Weighted {B}oolean Formulas},
  journal = {Theory and Practice of Logic Programming},
  volume  = {15},
  number  = {3},
  pages   = {358--401},
  year    = {2015},
}

@inproceedings{deepproblog,
  author    = {Manhaeve, Robin and Dumancic, Sebastijan and Kimmig, Angelika and Demeester, Thomas and {De Raedt}, Luc},
  title     = {{DeepProbLog}: Neural Probabilistic Logic Programming},
  booktitle = {Advances in Neural Information Processing Systems (NeurIPS)},
  year      = {2018},
}

@inproceedings{neurasp,
  author    = {Yang, Zhun and Ishay, Adam and Lee, Joohyung},
  title     = {{NeurASP}: Embracing Neural Networks into Answer Set Programming},
  booktitle = {Proceedings of the Twenty-Ninth International Joint Conference on Artificial Intelligence (IJCAI)},
  year      = {2020},
}

@inproceedings{parafrost,
  author    = {Osama, Muhammad and Wijs, Anton and Biere, Armin},
  title     = {{SAT} Solving with {GPU} Accelerated Inprocessing},
  booktitle = {Tools and Algorithms for the Construction and Analysis of Systems (TACAS)},
  series    = {Lecture Notes in Computer Science},
  volume    = {12651},
  pages     = {133--151},
  publisher = {Springer},
  year      = {2021},
  doi       = {10.1007/978-3-030-72016-2_8},
}

@article{dilp,
  author  = {Evans, Richard and Grefenstette, Edward},
  title   = {Learning Explanatory Rules from Noisy Data},
  journal = {Journal of Artificial Intelligence Research},
  volume  = {61},
  pages   = {1--64},
  year    = {2018},
}

@article{oled,
  author        = {Katzouris, Nikos and Artikis, Alexander and Paliouras, Georgios},
  title         = {Online Learning of {E}vent {D}efinitions},
  journal       = {Theory and Practice of Logic Programming},
  year          = {2016},
  eprint        = {1608.00100},
  archiveprefix = {arXiv},
}

@article{woled,
  author        = {Katzouris, Nikos and Paliouras, Georgios and Artikis, Alexander},
  title         = {Online Learning Probabilistic {E}vent {C}alculus Theories in {A}nswer {S}et {P}rogramming},
  journal       = {Theory and Practice of Logic Programming},
  volume        = {23},
  number        = {2},
  pages         = {362--386},
  year          = {2023},
  doi           = {10.1017/S1471068421000107},
  eprint        = {2104.00158},
  archiveprefix = {arXiv},
}

@article{brestais,
  author  = {Ray, Cyril and Dr{\'e}o, Richard and Camossi, Elena and Jousselme, Anne-Laure and Iphar, Cl{\'e}ment},
  title   = {Heterogeneous Integrated Dataset for Maritime Intelligence, Surveillance, and Reconnaissance},
  journal = {Data in Brief},
  volume  = {25},
  pages   = {104141},
  year    = {2019},
  doi     = {10.1016/j.dib.2019.104141},
}

@inproceedings{maritimecer,
  author    = {Pitsikalis, Manolis and Artikis, Alexander and Dr{\'e}o, Richard and Ray, Cyril and Camossi, Elena and Jousselme, Anne-Laure},
  title     = {Composite Event Recognition for Maritime Monitoring},
  booktitle = {Proceedings of the 13th ACM International Conference on Distributed and Event-based Systems (DEBS)},
  pages     = {163--174},
  year      = {2019},
  doi       = {10.1145/3328905.3329762},
}

@article{rtec,
  author  = {Artikis, Alexander and Sergot, Marek and Paliouras, Georgios},
  title   = {An {E}vent {C}alculus for Event Recognition},
  journal = {IEEE Transactions on Knowledge and Data Engineering},
  volume  = {27},
  number  = {4},
  pages   = {895--908},
  year    = {2015},
  doi     = {10.1109/TKDE.2014.2356476},
}

@inproceedings{d4compiler,
  author    = {Lagniez, Jean-Marie and Marquis, Pierre},
  title     = {An Improved Decision-{DNNF} Compiler},
  booktitle = {Proceedings of the Twenty-Sixth International Joint Conference on Artificial Intelligence (IJCAI)},
  year      = {2017},
}

@misc{dlpack,
  title        = {{DLPack}: Open In Memory Tensor Structure},
  howpublished = {\url{https://github.com/dmlc/dlpack}},
  note         = {Accessed 2026},
}

@misc{arrow,
  title        = {{Apache Arrow}: Cross-Language Development Platform for In-Memory Data},
  howpublished = {\url{https://arrow.apache.org}},
  note         = {Accessed 2026},
}

@inproceedings{c2d,
  author    = {Darwiche, Adnan},
  title     = {New Advances in Compiling {CNF} into Decomposable Negation Normal Form},
  booktitle = {Proceedings of the 16th European Conference on Artificial Intelligence (ECAI)},
  year      = {2004},
}

@article{tseitin,
  author  = {Tseitin, Grigori S.},
  title   = {On the Complexity of Derivation in Propositional Calculus},
  journal = {Studies in Constructive Mathematics and Mathematical Logic},
  year    = {1968},
}

@misc{fastfouriersat,
  author = {Cen, Yunuo and Zhang, Zhiwei and Fong, Xuanyao},
  title  = {Massively Parallel Continuous Local Search for Hybrid {SAT} Solving on {GPU}s},
  note   = {arXiv:2308.15020},
  year   = {2023},
}

@misc{pyo3,
  title        = {{PyO3}: Rust Bindings for {P}ython},
  howpublished = {\url{https://pyo3.rs}},
  note         = {Accessed 2026},
}

@misc{cudarc,
  title        = {cudarc: Safe Rust Wrapper around {CUDA}},
  howpublished = {\url{https://github.com/coreylowman/cudarc}},
  note         = {Accessed 2026},
}

@article{swiprolog,
  author    = {Wielemaker, Jan and Schrijvers, Tom and Triska, Markus and Lager, Torbj{\"o}rn},
  title     = {{SWI-Prolog}},
  journal   = {Theory and Practice of Logic Programming},
  volume    = {12},
  number    = {1-2},
  pages     = {67--96},
  year      = {2012},
  publisher = {Cambridge University Press},
}

@article{mercury,
  author  = {Somogyi, Zoltan and Henderson, Fergus and Conway, Thomas},
  title   = {The Execution Algorithm of {Mercury}, an Efficient Purely Declarative Logic Programming Language},
  journal = {Journal of Logic Programming},
  volume  = {29},
  number  = {1-3},
  pages   = {17--64},
  year    = {1996},
}

@inproceedings{souffle,
  author    = {Jordan, Herbert and Scholz, Bernhard and Suboti{\'c}, Pavle},
  title     = {{Souffl{\'e}}: On Synthesis of Program Analyzers},
  booktitle = {Computer Aided Verification (CAV)},
  pages     = {422--430},
  year      = {2016},
  publisher = {Springer},
}

@article{clingo,
  author  = {Gebser, Martin and Kaminski, Roland and Kaufmann, Benjamin and Schaub, Torsten},
  title   = {Multi-shot {ASP} Solving with {clingo}},
  journal = {Theory and Practice of Logic Programming},
  volume  = {19},
  number  = {1},
  pages   = {27--82},
  year    = {2019},
}

@article{ngo2018wcoj,
  author  = {Ngo, Hung Q. and Porat, Ely and R{\'e}, Christopher and Rudra, Atri},
  title   = {Worst-case Optimal Join Algorithms},
  journal = {Journal of the ACM},
  volume  = {65},
  number  = {3},
  pages   = {16:1--16:40},
  year    = {2018},
}

@inproceedings{veldhuizen2014leapfrog,
  author    = {Veldhuizen, Todd L.},
  title     = {Leapfrog Triejoin: A Simple, Worst-Case Optimal Join Algorithm},
  booktitle = {International Conference on Database Theory (ICDT)},
  pages     = {96--106},
  year      = {2014},
}

@article{wang2023freejoin,
  author  = {Wang, Yisu Remy and Willsey, Max and Suciu, Dan},
  title   = {Free Join: Unifying Worst-Case Optimal and Traditional Joins},
  journal = {Proceedings of the ACM on Management of Data (SIGMOD)},
  volume  = {1},
  number  = {2},
  pages   = {150:1--150:23},
  year    = {2023},
}

@phdthesis{andersen1994,
  author = {Andersen, Lars Ole},
  title  = {Program Analysis and Specialization for the {C} Programming Language},
  school = {University of Copenhagen},
  year   = {1994},
}

@inproceedings{ddisasm,
  author    = {Flores-Montoya, Antonio and Schulte, Eric},
  title     = {Datalog Disassembly},
  booktitle = {USENIX Security Symposium},
  pages     = {1075--1092},
  year      = {2020},
}

@inproceedings{lobster,
  author    = {Biberstein, Paul and Li, Ziyang and Devietti, Joseph and Naik, Mayur},
  title     = {{Lobster}: A {GPU}-Accelerated Framework for Neurosymbolic Programming},
  booktitle = {Architectural Support for Programming Languages and Operating Systems (ASPLOS)},
  year      = {2026},
  note      = {arXiv:2503.21937},
}

@article{scallop,
  author  = {Li, Ziyang and Huang, Jiani and Naik, Mayur},
  title   = {{Scallop}: A Language for Neurosymbolic Programming},
  journal = {Proceedings of the ACM on Programming Languages (PLDI)},
  volume  = {7},
  pages   = {166:1--166:25},
  year    = {2023},
}

@misc{sungpuwcoj,
  author       = {Sun, Yihao and Qi, Kunting and Gilray, Thomas and Kumar, Sidharth and Micinski, Kristopher},
  title        = {Scaling Worst-Case Optimal {Datalog} to {GPUs}},
  howpublished = {arXiv:2604.20073},
  year         = {2026},
}

@article{darwiche2001dnnf,
  author  = {Darwiche, Adnan},
  title   = {Decomposable Negation Normal Form},
  journal = {Journal of the ACM},
  volume  = {48},
  number  = {4},
  pages   = {608--647},
  year    = {2001},
}

@article{darwicheMarquis2002,
  author  = {Darwiche, Adnan and Marquis, Pierre},
  title   = {A Knowledge Compilation Map},
  journal = {Journal of Artificial Intelligence Research},
  volume  = {17},
  pages   = {229--264},
  year    = {2002},
}

@article{chaviraDarwiche2008,
  author  = {Chavira, Mark and Darwiche, Adnan},
  title   = {On Probabilistic Inference by Weighted Model Counting},
  journal = {Artificial Intelligence},
  volume  = {172},
  number  = {6-7},
  pages   = {772--799},
  year    = {2008},
}

@inproceedings{green2007provenance,
  author    = {Green, Todd J. and Karvounarakis, Grigoris and Tannen, Val},
  title     = {Provenance Semirings},
  booktitle = {Proceedings of the 26th ACM SIGMOD-SIGACT-SIGART Symposium on Principles of Database Systems (PODS)},
  pages     = {31--40},
  year      = {2007},
}

@inproceedings{chaff,
  author    = {Moskewicz, Matthew W. and Madigan, Conor F. and Zhao, Ying and Zhang, Lintao and Malik, Sharad},
  title     = {{Chaff}: Engineering an Efficient {SAT} Solver},
  booktitle = {Proceedings of the 38th Design Automation Conference (DAC)},
  pages     = {530--535},
  year      = {2001},
}

@inproceedings{sdd,
  author    = {Darwiche, Adnan},
  title     = {{SDD}: A New Canonical Representation of Propositional Knowledge Bases},
  booktitle = {Proceedings of the Twenty-Second International Joint Conference on Artificial Intelligence (IJCAI)},
  pages     = {819--826},
  year      = {2011},
}

@inproceedings{capelli2019,
  author    = {Capelli, Florent},
  title     = {Knowledge Compilation Languages as Proof Systems},
  booktitle = {Theory and Applications of Satisfiability Testing (SAT)},
  series    = {Lecture Notes in Computer Science},
  volume    = {11628},
  pages     = {90--99},
  year      = {2019},
  publisher = {Springer},
}

@inproceedings{satnet,
  author    = {Wang, Po-Wei and Donti, Priya L. and Wilder, Bryan and Kolter, J. Zico},
  title     = {{SATNet}: Bridging Deep Learning and Logical Reasoning Using a Differentiable Satisfiability Solver},
  booktitle = {Proceedings of the 36th International Conference on Machine Learning (ICML)},
  pages     = {6545--6554},
  year      = {2019},
}

@inproceedings{semanticloss,
  author    = {Xu, Jingyi and Zhang, Zilu and Friedman, Tal and Liang, Yitao and {Van den Broeck}, Guy},
  title     = {A Semantic Loss Function for Deep Learning with Symbolic Knowledge},
  booktitle = {Proceedings of the 35th International Conference on Machine Learning (ICML)},
  pages     = {5502--5511},
  year      = {2018},
}

@article{ltn,
  author  = {Badreddine, Samy and d'Avila Garcez, Artur and Serafini, Luciano and Spranger, Michael},
  title   = {Logic Tensor Networks},
  journal = {Artificial Intelligence},
  volume  = {303},
  pages   = {103649},
  year    = {2022},
}

\end{document}